%% file: root_C02.tex
\documentclass[twocolumn]{rsos}
\usepackage [a4paper,left=2.5cm,right=2cm,top=2.5cm, bottom=1.5cm]{geometry}
\usepackage{floatrow}
\usepackage[usenames,dvipsnames]{xcolor}

\usepackage{gensymb,mathtools}
\usepackage{graphicx,subcaption, adjustbox}
\usepackage{amsmath,amssymb,amsfonts,mathtools}
\usepackage{multicol,chngcntr, paralist,xspace}
\usepackage{multirow}
\usepackage{colortbl} 
\usepackage{relsize,scalerel,xspace} 
\usepackage{xargs,changepage,ifthen}
\usepackage{stfloats} 
\usepackage{nccmath} 
\usepackage{anyfontsize} 

\renewcommand{\thesubsection}{\thesection.\arabic{subsection}}
\renewcommand{\thesubsubsection}{\thesubsection.\arabic{subsubsection}}

\usepackage{titlesec}
\titlespacing*{\subsubsection}{0pt}{1.1\baselineskip}{\baselineskip}

\usepackage{siunitx}
\DeclareSIUnit\bw{BW}

\newcommand{\minus}{\scalebox{0.65}[1.0]{$-$}}   
\newcommand\myeq{\mkern1.5mu{=}\mkern1.5mu}
\newcommand\mycross{\mkern1.5mu{\times}\mkern1.5mu}
\newcommand\VPa[1][color_blue]{$\mathrm{VP_{\color{#1}{A}}}$\xspace}
\newcommand\VPb[1][color_red]{$\mathrm{VP_{\color{#1}{B}}}$\xspace}
\newcommand\VPbl[1][color_darkred]{$\mathrm{VP_{\color{#1}{BL}}}$\xspace}

\usepackage[numbers,sort&compress,square,comma]{natbib} 

\makeatletter
\renewcommand\@biblabel[1]{#1.}
\makeatother

\makeatletter
\def\NAT@def@citea{\def\@citea{\NAT@separator\,}}
\makeatother

 \newenvironment{noDisplaySkip}{\setlength{\abovedisplayskip}{7pt}\setlength{\belowdisplayskip}{-1.5pt}}{}

\makeatletter
\patchcmd{\thebibliography}{\c@NAT@ctr\z@}{\c@NAT@ctr\z@
  \setlength{\labelwidth}{1.2em}%
  \setlength{\labelsep}{.5em}%
  \setlength{\leftmargin}{\dimexpr\labelwidth+\labelsep}%
}{}{}
\makeatother

\usepackage{nomencl}
\makenomenclature
\renewcommand\nomgroup[1]{%
  \item[\bfseries
  \ifstrequal{#1}{A}{General terminology}{%
  \ifstrequal{#1}{B}{Symbols related to the experiment}{%
  \ifstrequal{#1}{C}{Symbols related to the simulation}{%
  \ifstrequal{#1}{D}{Superscripts}{%
  \ifstrequal{#1}{E}{Subscripts}{}}}}}%
]}

\usepackage[capitalize,noabbrev,nameinlink]{cleveref}
\newcommand{\cmmnt}[1]{\ignorespaces}
\creflabelformat{equation}{#2\textup{#1}#3}

\newcommand{\creff}[2]{figure~\labelcref{#1}{#2}} 
\newcommand{\creft}[2]{table~\labelcref{#1}{#2}} 
\newcommand{\crefe}[1]{equation~\labelcref{#1}} 

\newcommand{\crefs}[5]{figures~\labelcref{#1}{#2}{#3}\labelcref{#4}{#5}}

\newcommand{\crefsss}[9]{figures~\labelcref{#1}{#2}{#3}\labelcref{#4}{#5}\labelcref{#4}{#6}{#7}\labelcref{#8}{#9}}

\newcommand{\crefsub}[3]{figure~\labelcref{#1}{#2}-\,\labelcref{#1}{#3}}

\usepackage[bottom]{footmisc}
\renewcommand{\footnotesize}{\fontsize{8pt}{9pt}\selectfont}

\usepackage{etoolbox}
\makeatletter
\newcommand{\mainmatter}{
  \setcounter{footnote}{0}%
  \let\@fnsymbol\@arabic
  \def\@makefnmark{\textsuperscript{\arabic{footnote}}}%
}
\makeatother

\usepackage{hypernat}
\usepackage{footnotehyper}
\hypersetup{colorlinks=false,urlcolor=black}

\makeatletter  
\def\mathcolor#1#{\@mathcolor{#1}}
\def\@mathcolor#1#2#3{%
  \protect\leavevmode
  \begingroup
    \color#1{#2}#3%
  \endgroup
}
\makeatother 

\definecolor{color_gray}{rgb}{0.5 0.5 0.5}
\definecolor{color_darkgray}{rgb}{0.3 0.3 0.3}
\definecolor{color_red}{rgb}{0.6980    0.0941    0.1686} 
\definecolor{color_darkred}{rgb}{0.3017    0.0004    0.0026} 
\definecolor{color_blue}{rgb}{0.1294    0.4000    0.6745}
\definecolor{color_darkblue}{rgb}{0.0011    0.0472    0.2691} %
\definecolor{color_yellow}{rgb}{1.0000    0.8871    0.5453}
\definecolor{color_teal}{rgb}{0.5804    0.8275    0.7765}

\usepackage{tikz}
\usetikzlibrary{shapes, calc, arrows, shadows, positioning,decorations.pathreplacing} 
\newenvironment {annotatedFigure}[1]{\centering\begin{tikzpicture}[remember picture]
\node[anchor=south west,inner sep=0] (image) at (0,0) {#1};\begin{scope}[x={(image.south east)},y={(image.north west)}]}{\end{scope}\end{tikzpicture}}

\newcommand*\sublabel[4]{
\node at (#2) [fill=none,shape=circle,draw=none, scale=1,inner sep=1pt,font=\sffamily,text=#3] {\text{\relscale{#4}{#1}}};}

\newcommandx\markerSquare[2][1=color_darkgray,2=0]{\raisebox{-0.5 pt}{\tikz{\node[draw=white,scale=0.8,shape=rectangle,fill={#1},rotate={#2}](){};}}}
\tikzset{cross/.style={cross out, draw, fill=none, minimum size=2*(#1-\pgflinewidth), inner sep=0pt, outer sep=0pt}, cross/.default={3.5pt}}
\newcommandx\markerCross[2][1=color_darkgray,2=1.1]{\raisebox{-0.2 pt}{\tikz{\node[draw={#1},scale={#2},very thick,cross,rotate=90](){};}}}
\newcommand\markerCircle[1][color_darkgray]{\raisebox{-0.5 pt}{\tikz{\node[draw=white,scale=0.6,circle,fill=#1,opacity=1](){};}}}
\newcommandx\markerRectangle[2][1=color_darkgray,2=0]{\raisebox{-0.3 pt}{\tikz{\node[draw=white,scale=0.8,shape=rectangle,fill={#1},rotate={#2},minimum width=1em, minimum height=0.5em](){};}}}
\newcommandx\markerDiamond[2][1=color_darkgray,2=0]{\raisebox{-2.5 pt}{\tikz{\node[draw=white,scale=0.5,shape=diamond,fill={#1},rotate={#2},aspect=0.5,minimum width=1em, minimum height=1em](){};}}}
\newcommand\markerLine[1][color_darkgray]{\raisebox{2 pt}{\tikz{\draw[line width=0.5mm, {#1}](0,0)--(0.3,0);}}}
\newcommand\markerDashedLine[1][color_darkgray]{\raisebox{2 pt}{\tikz{\draw[dashed, line width=0.5mm, {#1}](0,0)--(0.3,0);}}}

\newcommand\circledRmark[1][color_gray]{\raisebox{-2 pt}
{\tikz\node[draw=#1,scale=0.8,circle,fill=none,inner sep=0.3pt]{$\scaleto{\, \mathsf{\color{#1}{R}} \,}{4pt}$\xspace};}}

\begin{document}{}

\input{sec/title.tex}

\onecolumn
\begin{abstract} %

\input{sec/abstract.tex}
\end{abstract} %
\twocolumn

\begin{fmtext}
\section{Introduction}\label{sec:intro}
\input{sec/intro_P1.tex}
\end{fmtext}

\maketitle
\clearpage

\input{sec/intro_P2.tex}

\section{ Methods}\label{sec:methods}
\subsection{Experimental Methods}\label{subsec:experimentalmethods}
\input{sec/method_exp.tex}

\subsection{Simulation Methods}\label{subsec:simulationmethods}
\input{sec/method_sim.tex}

\section{ Results}\label{sec:results}
\subsection{Experimental Results}\label{subsec:experimentalresults}
\input{sec/result_exp.tex}


\subsection{Simulation Results }\label{subsec:simulationresults}
\input{sec/result_sim.tex}

\section{Discussion}\label{sec:discussion}
\input{sec/discussion.tex}

\section{Conclusion}\label{sec:conclusion}
\input{sec/conclusion.tex}

\section{Acknowledgement}\label{sec:acknowledgement}
\input{sec/acknowledgement.tex}

\section{Data availability}\label{sec:Data_availability}
\input{sec/data_availability.tex}


\input{sec/nomenclature.tex}

\clearpage\newpage
\bibliographystyle{apalike} 
{
\renewcommand{\clearpage}{}  
\onecolumn
\setlength{\columnsep}{0.5cm}
\begin{multicols*}{3}
\bibliography{literaturRunning}
\end{multicols*}
\twocolumn
}

\clearpage \newpage
\appendix
\section{Appendix}\label{sec:appendix}
\renewcommand{\thefigure}{A\arabic{figure}}
\renewcommand{\thetable}{A\arabic{table}}
\setcounter{figure}{0}
\setcounter{table}{0}
\input{sec/appendix.tex}

\end{document}

%% file: sec/title.tex

\title{Postural Stability in Human Running with Step-down Perturbations \\ {\Large An Experimental and Numerical Study}}

\author{%
\fontsize{14}{21}\selectfont {\"O}zge Drama$^{1,\dagger}$,  Johanna Vielemeyer$^{2,3,\dagger}$, \\ Alexander Badri-Spr{\"o}witz$^{1}$, and Roy M{\"u}ller$^{2,3}$}

\address{\fontsize{9}{13}\selectfont{$^{1}$Dynamic Locomotion Group, Max Planck Institute for Intelligent Systems, Germany \\
$^{2}$Department of Neurology/ Orthopedic Surgery, Klinikum Bayreuth GmbH, Germany\\
$^{3}$Department of Motion Science, Friedrich Schiller University-Jena, Germany \\
}}

\subject{Biomechanics}

\keywords{Bipedal locomotion, Human running, Step-down perturbation, Postural stability, TSLIP model, Virtual point (VP, VPP)}

\corres{{\"O}zge Drama\\
\email{drama@is.mpg.de}}
\others{$\dagger$ Shared first authorship. J.V. performed the gait analysis for human running. {\"O}.D. shares first authorship due to generation of the simulation model and analysis.  Both J.V. and {\"O}.D. wrote the manuscript and all authors discussed the results and contributed to the final manuscript. \vspace{3cm}}


%% file: sec/abstract.tex
Postural stability is one of the most crucial elements in bipedal locomotion.
Bipeds are dynamically unstable and need to maintain their trunk upright against the rotations induced by the ground reaction forces (GRFs), especially when running. 
Gait studies report that the GRF vectors focus around a virtual point above the center of mass (\VPa), while the trunk moves forward in pitch axis during the stance phase of human running. 
However, a recent simulation study suggests that a virtual point below the center of mass (\VPb) might be present in human running, since a \VPa yields backward trunk rotation during the stance phase. 
In this work, we perform a gait analysis to investigate the existence and location of the VP in human running at \SI{5}{\meter\per\second}, and support our findings numerically using the spring-loaded inverted pendulum model with a trunk (TSLIP).
We extend our analysis to include perturbations in terrain height (visible and camouflaged), and investigate the response of the VP mechanism to step-down perturbations both experimentally and numerically.
Our experimental results show that the human running gait displays a \VPb of \SI{\approx \minus 30}{\centi\meter} and a forward trunk motion during the stance phase.
The camouflaged step-down perturbations affect the location of the \VPb. 
Our simulation results suggest that the \VPb is able to encounter the step-down perturbations and bring the system back to its initial equilibrium state.

%% file: sec/intro_P1.tex
Bipedal locomotion in humans poses challenges for stabilizing the upright body due to the under-actuation of the trunk and the hybrid dynamics of the bipedal structure.

Human gait studies investigate the underlying mechanisms to achieve and maintain the postural stability in symmetrical gaits such as walking and running.
One major observation states that the ground reaction forces (GRFs) intersect near a virtual point (VP) above the center of mass (CoM) \cite{maus2010upright}.
Subsequent gait studies report that the VP is \SI{15 \minus 50}{\centi\meter} above the CoM (\VPa) in sagittal plane for level walking \cite{gruben2012force,maus2010upright,muller2017force,Sharbafi_2015a,vielemeyer2019ground}.
Among those, only a single study reports a limited set of level walking trials with a VP below the CoM (\VPb) \cite{maus2010upright}.
%
%
The \VPa strategy is also observed when coping with the step-down perturbations in human walking, even when walking down a camouflaged curb \cite{vielemeyer2019ground}.
A similar behavior is observed for the avians, where a \VPa of~\SI{5}{\centi\meter} is reported for level walking, grounded running, and running of the quail \cite{andrada2014trunk,blickhan2015positioning}.
%
Unlike in the studies with healthy subjects, it is reported that humans with Parkinson's disease display a \VPb when walking \cite{Scholl_2018}.
In addition, a \VPb was identified in the frontal plane for human level walking \cite{Firouzi_2019}.
The existing literature for human running report a \VPa \cite{blickhan2015positioning,Maus_1982}. However, these experiments are limited to a small subset of subjects and trials, hence are not conclusive.

%% file: sec/intro_P2.tex
The observation of the GRFs intersecting at a virtual point suggests that
there is potentially a control mechanism to regulate the whole body angular momentum \cite{maus2008stable,maus2010upright,Hinrichs_1987}.
Based on this premise, the behavior of a VP based postural mechanism would depend of the location and adjustment of the VP.
%
It also raises the question whether the VP position depends on the gait type, locomotor task (e.g., control intent) and terrain conditions.
%
%

The spring-mass model (SLIP) is extensively used in gait analysis due to its capability to reproduce the key features of bipedal locomotion. The SLIP model is able to reproduce the CoM dynamics observed in human walking \cite{geyer2006compliant} and running \cite{blickhan1989spring,McMahon_1990,Mueller_2016}.
%
This model can be extended with a rigid body (TSLIP) to incorporate the inertial effects of an under-actuated trunk, where the trunk is stabilized through a torque applied at the hip \cite{Maus_1982,maus2008stable,maus2010upright}.

Based on the experimental observations, the VP is proposed as a control method to determine the hip torque in the TSLIP model to achieve postural stability \cite{maus2008stable}.
The VP as a control mechanism in TSLIP model is implemented for human walking \cite{Lee_2017_I,Maufroy_2011,Sharbafi_2015a,Vu_2017_III,Vu_2017_II}, hopping \cite{Sharbafi_2012,Sharbafi_2013}, running \cite{maus2008stable,drama2019human,VanBommel_2011}, and avian gaits \cite{andrada2014trunk,drama2019bird}.
%
It is also implemented and tested on the ATRIAS robot for a walking gait \cite{Peekema_2015}.
Yet the currently deployed robotic studies are limited to a small set of gait properties (e.g., forward speed) and simple level terrain conditions.
%

In the simulation model, the selection of the VP position influences the energetics of the system by distributing the work performed by the leg and the hip \cite{drama2019human,drama2019bird}. A \VPb in the human TSLIP model reduces the leg loading at the cost of increased peak hip torques for steady-state gaits. A \VPa yields lower duty factors and hence higher peak vertical GRF magnitudes, whereas a \VPb yields larger peak horizontal GRF magnitudes. Consequently, a \VPa can be used to reduce the kinetic energy fluctuations of the CoM, and a \VPb to reduce the potential energy fluctuations.

In human gait, the trunk moves forward during the single stance phase of walking and running, which is reversed by a backward trunk motion in double stance phase of walking \cite{thorstensson1984trunk} and flight phase of running \cite{Maus_1982,thorstensson1984trunk}.
In TSLIP model simulations of human running, the trunk moves forward during the stance phase if a \VPb is used, whereas it moves backward for a \VPa \cite{drama2019bird,drama2019human,maus2008stable,sharbafi2014stable}.

One potential reason for the differences between the human and the model may be that the TSLIP model does not distinguish between the trunk and whole body dynamics.
In human walking, the trunk pitching motion is reported to be \SI{180}{\degree} out-of-phase with the whole body \cite{gruben2012force}.
A \VPa in the TSLIP model predicts the whole body dynamics with backward rotation, and it follows that the trunk rotation is in the opposite direction (i.e., forward).
The phase relation between the trunk and whole body rotation has not been published for human running, to our knowledge.
However, we can indirectly deduce this relation from the pitch angular momentum patterns.
In human running, the pitch angular momentum of the trunk and the whole body are inphase, and they both become negative during stance phase (i.e., clockwise rotation of the runner) \cite{Hinrichs_1987}.
The negative angular momentum indicates that the GRFs should pass below the CoM. Therefore, a \VPb in the TSLIP model is able to predict the whole body dynamics with forward rotation, and the trunk rotation is in the same direction (i.e., forward).

The VP can also be used to maneuver, when the VP target is placed out of the trunk axis  \cite{maus2008stable,Sharbafi_2013}.
A simulation study proposes to shift the VP position horizontally as a mechanism to handle stairs and slopes \cite{Kenwright_2011}.
%
The gait analyses provide insights into the responses of GRFs to changes in terrain.
In human running, step-down perturbations increase the magnitude of the peak vertical GRF.
The increase gets even higher if the drop is camouflaged \citep{muller2012leg}.
However, there is no formalism to describe how the VP position relates to the increase in GRFs in handing varying terrain conditions.

In the first part of our work, we perform an experimental analysis to acquire trunk motion patterns and ground reaction force characteristics during human running.
Our gait analysis involves human level running, and running over visible and camouflaged step-down perturbations of \SI{\minus 10}{\centi\meter}.
We expect to observe a net forward trunk pitch motion during the stance phase of running based on the observation in \cite{thorstensson1984trunk}, and estimate a \VPb from the GRF data based on the hypothesis in~\cite{drama2019human}.

In the second part, we perform a simulation analysis using the TSLIP model with the gait parameters estimated from our experiments.
We generate an initial set of gaits that match to the experimental setup, and extend our analysis to larger set of step-down perturbations up to \SI{\minus 40}{\centi\meter}, which is close to the maximum achievable perturbation magnitude in avians \cite{BirnJeffery_2014}.
We investigate whether a \VPb controller is able to stabilize the gait against the step-down perturbations, and if so, how does it contribute to the energy flow in counteracting the perturbation.

%% file: sec/method_exp.tex
In this section, we describe the experimental setup and measurement methods.
In our experiments, ten physically active volunteers (9 male, 1 female, mean $\pm$ s.d., age: {$24.1 \pm 3.4$} years, mass: \SI{73.8 \pm 7.3}{\kilo\gram}, height: \SI{179.9 \pm 7.6}{\centi\meter}) are instructed to run over a \SI{17}{\meter} track.
The running track has two consecutive force plates in its center, where
the first plate is fixed at ground-level, and the second one is height adjustable.
%
We designed three sets of experiments, where the subjects were asked to run at their self-selected velocity\footnote{The velocity was calculated for the stance phases of both contacts.} (\SI{4.9 \pm 0.5}{\meter\per\second}, \creft{tab:VP_statistical}{}).
In the first experiment, the subjects were asked to run on a track with an even ground (V0).
In the second experiment, the second force plate was lowered \SI{\minus 10}{\centi\meter}, which was visible to the subjects (V10).
In the third experiment, the second force plate was lowered \SI{\minus 10}{\centi\meter}, and an opaque sheet was added on top of the plate on ground level to camouflage the drop. A wooden block was randomly placed between the second force plate and the opaque sheet during the course of the experiment without subject's knowledge. In other words, the subjects were not aware whether the step would be on the ground level (C0), or would be a step-down drop (C10).
The step corresponding onto the first force plate is referred to as step~-1, and the step to the second force plate as step~0.

\begin{figure}[t!]
{\captionof{figure} {Experimental setup. The first force plate is on the ground level, whereas the second force plate is height adjustable (step~0).
The camouflaged setting for the second force plate is shown on right for elevations of \SI{0}{\centi\meter} (C0, blue) and \SI{-10}{\centi\meter} (C10, red).
The placement of the motion capture markers is given on left, where the markers are denoted the letters A-G.
The trunk angle is shown with $\gamma$ and is positive in the counterclockwise direction.
}
\label{fig:Figure_Exp_setup}}
{\begin{annotatedFigure}
  {\includegraphics[width=1\linewidth]{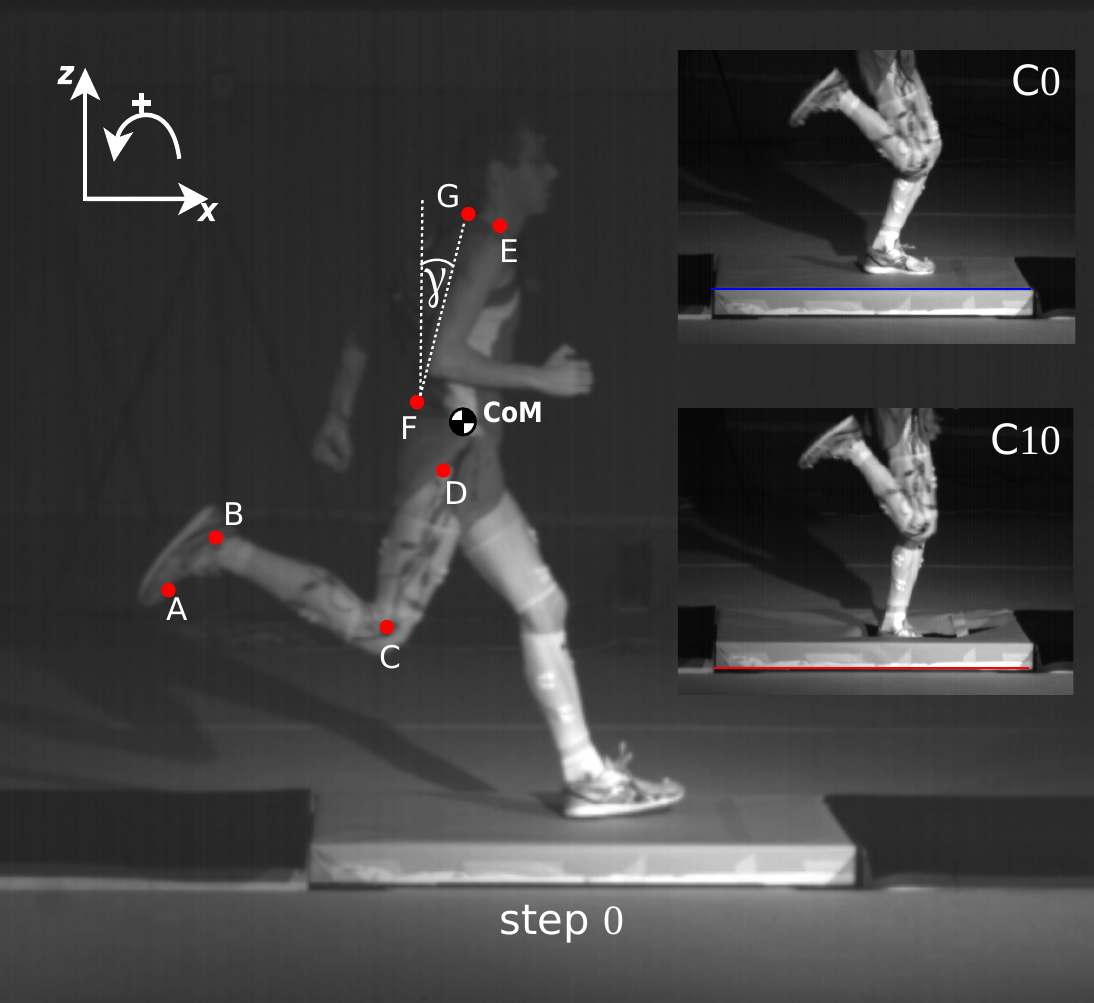}}
\end{annotatedFigure}
}
\end{figure}

All trials were recorded with eight cameras by a 3D motion capture system working with infrared light.
In sum twelve spherical reflective joint markers (19 mm diameter) were placed on the tip of the fifth toe [A], malleolus lateralis [B], epicondylus lateralis femoris [C], trochanter major [D], and acromion [E] on both sides of the body as well as on L5 [F] and C7 [G] processus spinosus (see \creff{fig:Figure_Exp_setup}{}).
The CoM was determined with a body segment parameter method according
to \citet{winter2009biomechanics}. 
The trunk angle $\gamma$ was calculated from the line joining C7 to L5 with respect to the vertical \cite{muller2014kinetic}.

Further information concerning the participants, and the technical details of the measurement equipment (i.e., force plates, cameras) can be found in \citet{muller2012leg} and partly in \citet{ernst2014vertical}.

The method for analyzing the gait data and estimating a potential VP is analogous to the gait analysis carried out for the human walking in \cite{vielemeyer2019ground}.
Here, we denote the intersection point of the GRF vectors as a VP without implications for this point being above or below the CoM.
%
To compute the VP, we use the instantaneous GRF vectors, which have an origin at the center of pressure (CoP) and are expressed in a CoM-centered coordinate frame that aligns with the gravity vector in vertical axis \cite{muller2017force}.
The CoP is calculated from the kinetic data using the method described in \citet{winter2009biomechanics}.
Then, the VP is estimated as the point, which minimizes the sum of the squared distances between the GRF vectors and itself.
%
For the camouflaged setting with a wooden block placed on the force plate (C0), we can not calculate the CoP accurately. Thus, the VP is not estimated for C0 case.
%

The human gait data involves impact forces at the leg touch-down, which introduces an additional behavior in the GRF pattern \cite{muller2012leg,gunther2003dealing,vanderLinden2009hitting}.
In order to see the influence of the impact on VP, we are presenting our recorded data in two ways. First calculation involves the full GRF data from leg touch-down to take-off (\SI{100}{\percent} dataset), whereas the second calculation involves the GRF data starting from \SI{10}{\percent} of the stance to the leg take-off (\SI{90}{\percent} dataset).
%

In theory of VP, all of the GRF vectors start from the CoP and point to a single virtual point. However, the human gait data differs from this theoretical case, as the human is more complex.
%
To evaluate the amount of agreement between the theoretical VP based forces and experimentally measured GRFs, we use a measure called the \emph{coefficient of determination} ($R^2$) similar to \citet{herr2008angular}:

\begin{equation}
\label{R_squared}
R^2 \; =\;  \left ( 1\; - \;\frac{\mathlarger{\mathlarger{\sum\limits}}_{i=1}^{N_{trial}} \mathlarger{\mathlarger{\sum}}\limits_{j=1}^{N_{\%}} \left( \theta^{ij}_{exp}-\theta^{ij}_{theo}\right)^2}{\mathlarger{\mathlarger{\sum}}\limits_{i=1}^{N_{trial}} \mathlarger{\mathlarger{\sum}}\limits_{j=1}^{N_{\%}} \left( \theta^{ij}_{exp}-\overline{\theta}_{exp}\right)^2} \right )\times 100\%,
\end{equation}
The ($\theta_{exp}$,\,$\theta_{theo}$) are the experimental GRF and theoretical force vector angles, $N_{trial}$ is the number of trials, and $N_{\%}=100$ is the measurement time. Here, $\overline{\theta}_{exp}$ is the mean of the experimental GRF angles over all trials and measurement times.
The number of trials is equal to 30 for visible conditions (15 for V0 and 15 for V10) and 20 for the camouflaged conditions (12 for C0 and 8 for C10).

Note that $R^2=\protect\SI{100}{\percent}$ if there is a perfect fit for the experimental GRF and the theoretical force vector angles. The value of $R^2$ aproaches zero as the estimation of the model is equal to the use of ${\theta}_{exp}$ as an estimator \cite{herr2008angular}.

We also compute the horizontal and vertical impulses $\vec{p}$ for two intervals (braking and propulsion) by integrating the GRFs over time.
The braking interval went from touch-down to mid-stance (zero-crossing of the horizontal GRFs) and the propulsion interval mid-stance onward.
We report the values for brake-propulsion intervals individually in \cref{subsec:experimentalresults}.
To enable the comparison among subjects, we normalize the impulses to each subject's body weight (\si{\bw}), leg length ($l$, the distance between lateral malleolus and trochanter major of the leg in contact with the ground) and standard gravity ($g$) in accordance with \cite{hof1996scaling} as,
\begin{equation}
\label{eqn:impulse}
\vec{p}_{normalized} \; =\;  \frac{\vec{p}}{\mathrm{BW} \cdot \sqrt{l/g}}.
\end{equation}

Because of the inaccuracy in calculating the CoP, we did not analyze the C0 statistically.
For all other experimental settings (V0, V10, and C10), we used repeated measures ANOVA ($p<0.05$) with post hoc analysis (\v{S}id\'{a}k correction) to test the statistical significance of the estimated VP position, the impulses and additional gait properties.
In order to verify whether the VP is above or below the CoM (\VPa or \VPb), we performed a one-sample t-test compared with zero, separately for each condition with  \v{S}id\'{a}k correction.

%% file: sec/method_sim.tex
In this section, we describe the TSLIP model that we use to analyze how the VP reacts to the step-down perturbations in human running. The TSLIP model consists of a trunk with mass $m$ and moment of inertia $J$, which is attached to a massless leg of length $l$ and a massless point foot $F$ (see \creff{fig:TSLIPmodel}{a}). The leg is passively compliant with a parallel spring-damper mechanism, whereas the hip is actuated with a torque ${\tau}_{H}$. The dynamics of the system is hybrid, which involves a flight phase that has ballistic motion, followed by a stance phase that reflects the dynamics of the spring-damper-hip mechanism. The phases switch when the foot becomes in contact with the ground at touch-down, and when the leg extends to its rest length $l_{0}$ at take-off.

The equations of motion for the CoM state $(x_{C},  z_{C},  \theta_{C}) $ during the stance phase can be written~as in \crefe{eqn:EoM}, where the linear leg spring force $F_{sp} \myeq k \, (l \minus l_{0})$ and bilinear leg damping force $F_{dp} \myeq c \, \dot{l} \, (l  \minus l_{0})$ generate the axial component of the GRF in foot frame $ \prescript{}{F}{\mathbf{F}}_{a}  \myeq \left(F_{sp} \minus F_{dp} \right) {\left[ \minus \cos\theta_{L} \; \sin\theta_{L}  \right]}^\mathsf{T}$. Here,~$k$ refers to the spring stiffness and $c$ to the damping coefficient. ~The hip~torque ${\tau}_{H}$  creates the tangential component of the GRF $ \prescript{}{F}{\mathbf{F}}_{t} \myeq \left( \scaleto{\frac{\minus \tau_{H}}{l_{L}}}{12pt} \right) {\left[ \sin\theta_{L} \;  \minus \cos\theta_{L}  \right] }^\mathsf{T},$ (see \creff{fig:TSLIPmodel}{d}).
%
{
{\noDisplaySkip
\begin{equation}
\begin{aligned}
m \begin{bmatrix}  \ddot{x}_{C} \\  \ddot{z}_{C} \end{bmatrix} & \myeq \prescript{}{F}{\mathbf{F}}_{a} +  \prescript{}{F}{\mathbf{F}}_{t} + g, \\ \vspace{0.5cm}
J\, \ddot{\theta}_{C} & \myeq \minus {\mathbf{r}}_{FC} \mycross (  \prescript{}{F}{\mathbf{F}}_{a} +  \prescript{}{F}{\mathbf{F}}_{t}).
\label{eqn:EoM}
\end{aligned}
\end{equation}}

\begin{figure}[t!]
{\captionof{figure}{a) TSLIP model that shows the forward (anterior) and backward (posterior) trunk motion. b) Vector notations used in equations of motion. c) The parameter space for the VP is divided into two regions: the virtual points above the center of mass (\VPa) and below (\VPb). \VPa causes backward and \VPb causes forward trunk rotation during the stance phase. Each subspace is divided further with respect to the leg axis, where the sign of the hip torque changes. d) For \VPb, the points above the leg axis yield a negative and points below (\VPbl) yield positive hip torque at touch-down. The VP is described with the radius ($r_{VP}$) and angle ($\theta_{VP}$) that is expressed in CoM centered world coordinate frame. Here presented human running experiments reveal that the VP is \SI{\protect\minus 30}{\centi\meter} below the CoM (see \cref{subsec:experimentalresults}). This corresponds to the \VPbl region with -\SI{180}{\degree} VP angle in our simulation.
}\label{fig:TSLIPmodel}}
{\begin{annotatedFigure}
	{\includegraphics[width=1\linewidth]{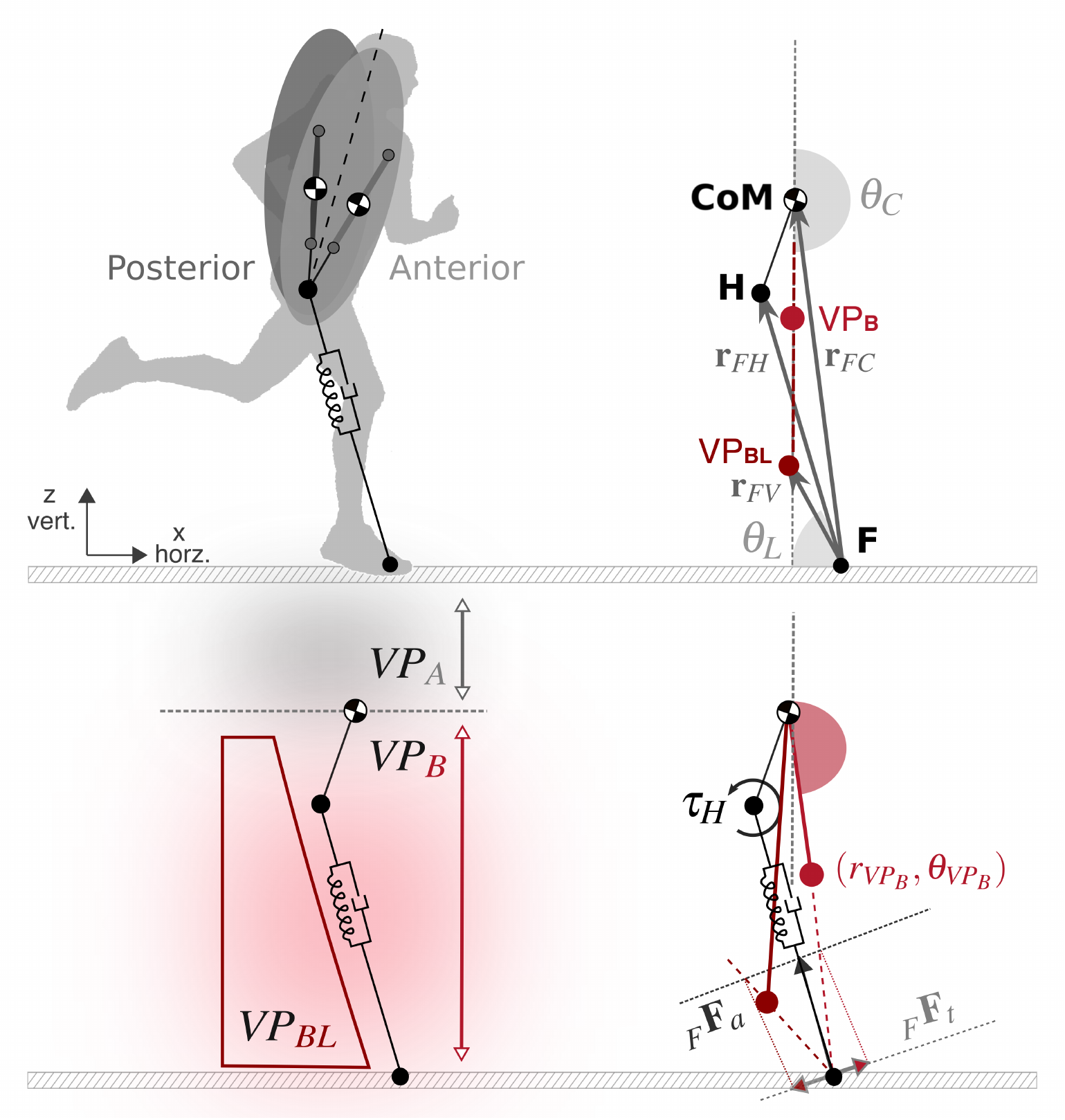}}
	\sublabel{a)}{0.08,0.945}{color_gray}{0.9}
	\sublabel{b)}{0.6,0.945}{color_gray}{0.7}
	\sublabel{c)}{0.08,0.43}{color_gray}{0.7}
	\sublabel{d)}{0.6,0.43}{color_gray}{0.7}
\end{annotatedFigure}}
\end{figure}

The leg and the hip maintain the energy balance of the system. The hip increases the system energy to propel the body forward, whereas the leg damper removes an equivalent energy in return. We determine $\tau_{H}$, such that the GRF points to a virtual point (VP), which is characterized by the radius $r_{VP}$ (i.e.,~distance between the hip and CoM) and angle $\theta_{VP}$, as shown in \creff{fig:TSLIPmodel}{d} ({\protect \markerCircle[color_red]}). The hip torque as a function of the VP is written~as,
%
{\noDisplaySkip
\begin{equation}
\begin{aligned}
\tau_{H} &= \tau_{VP} =  \prescript{}{F}{\mathbf{F}}_{a} \times  \left[   \frac{\mathbf{r}_{FV} \times \mathbf{r}_{FH} }{\mathbf{r}_{FV} \cdot \mathbf{r}_{FH}}\right]   \times  l,  \\  \vspace{0.5cm}
\mathbf{r}_{FV} &= \mathbf{r}_{FC}  + r_{VP}   \begin{bmatrix*}[r] \minus \sin \left( \theta_{C}+\theta_{VP} \right) \\ \cos \left( \theta_{C}+\theta_{VP} \right) \end{bmatrix*}.
\end{aligned}
\label{eqn:tauVP}
\end{equation}}
%

We utilize two linear controllers: one for the leg angle at touch-down $ \theta_{L}^{TD}$, and other for the VP angle $\theta_{VP}$, both of which are executed at the beginning of the step at apex, as shown in \creff{fig:Flowchart}{}.
The leg angle is regulated as,
{\noDisplaySkip
\begin{equation}  \label{eqn:thetaL}
\begin{adjustbox}{width=0.85\columnwidth}
$\begin{aligned}
 \theta_{L}^{TD} \, |_{i} =  \theta_{L}^{TD} \, |_{i \minus 1} +
 k_{\dot{x}_{0}}    (  \Delta \dot{x}_{C}^{AP} \, |_{\minus 1}^{i} )
 +  k_{\dot{x}}       (  \Delta \dot{x}_{C}^{AP} \, |_{i  \minus 1}^{i} ),
   \end{aligned}$
\end{adjustbox}
\end{equation}}

\noindent with $\Delta \dot{x} |_{\text{-}1}^{i}$ being the difference in apex velocity $\dot{x}$ between time steps -$1$ and~$i$.
The VP angle is defined with respect to a CoM-centered, stationary coordinate frame that is aligned with the global vertical axis, if the VP is set below the CoM (see \crefs{fig:TSLIPmodel}{b}{,\,}{fig:TSLIPmodel}{d}) \cite{drama2019bird}}.
It is adjusted based on the difference between the desired mean body angle $ \theta_{C}^{Des}$, and the mean body angle observed in the last step $\Delta\theta_{C}$~as,
{\noDisplaySkip
\begin{equation} \label{eqn:thetaVP}
\begin{aligned}
\theta_{VP} \ |_{i} =\theta_{VP} \ |_{i-1} + k_{vp} \left( \theta_{C}^{Des} \minus \Delta\theta_{C}   \right ).
\end{aligned}
\end{equation}}
%

The model parameters are selected to match a \SI{80}{\kilo\gram} human with \SI{1}{\meter} leg length (see \creft{tab:ModelPrm}{} for details).
%
The damping coefficient is set to $c \protect \myeq$\protect\SI{680}{\kilo\newton\second\per\meter} to match the trunk angular excursion of \protect\SI{4.5}{\degree} reported in \cite{Heitcamp_2012,Schache_1999,thorstensson1984trunk}.
%
The forward speed and VP radius are set to \SI{5}{\meter\per\second} and \SI{\minus 30}{\centi\meter} respectively, to match our estimated gait data in \creft{tab:VP_statistical}{}.
A VP radius of \SI{\minus 30}{\centi\meter} becomes below the leg axis at leg touch-down with the model parameters we chose.
Since the position of VP relative to the leg axis affects the sign of the hip torque, the \VPb region is separated into two and the points below the leg axis are called \VPbl (\crefsub{fig:TSLIPmodel}{c}{d}), in accordance with \cite{drama2019human}.
%

First, we generate a {\color{color_gray}{base gait}} for level running using the framework in \cite{drama2019human}, which corresponds to the V0 in our human running experiments.
%
Then, we introduce step-down perturbations of $\Delta z \myeq$[\SI{ \minus 10, \minus 20,  \minus 30, \minus 40}{\centi\meter}] in step~0. The \SI{ \minus 10}{\centi\meter} drop corresponds to the V10 and C10 of the human running experiments.
In the simulations, the VP controller is blind to the changes in step~0, since the controller update happens only at the apex of each step. The~postural correction starts at step~1. In other words, we see~the natural response of the system at step~0, and counteracting response of the VP controller starting from step~1. In contrast, there might already be a postural control during the step~0 in the experiments.

The step-down perturbation increases the total energy of the system. The added energy can be either dissipated e.g., via the hip torque or leg damper, or converted to other forms of energy e.g., change in speed or hopping height. In the latter case, we need to update the desired forward speed in the leg angle control (\crefe{eqn:thetaL}) until all excess energy is converted to kinetic energy.

We implemented the TSLIP model in Matlab\textsuperscript{\circledRmark} using variable step solver ode113 with a relative and absolute integrator error tolerance of $1 \times 10^{\minus 12}$.

%% file: sec/result_exp.tex

\begin{figure}[!ht]
  \begin{floatrow}
  \ffigbox[\Xhsize]
  {\includegraphics[width=0.97\linewidth]{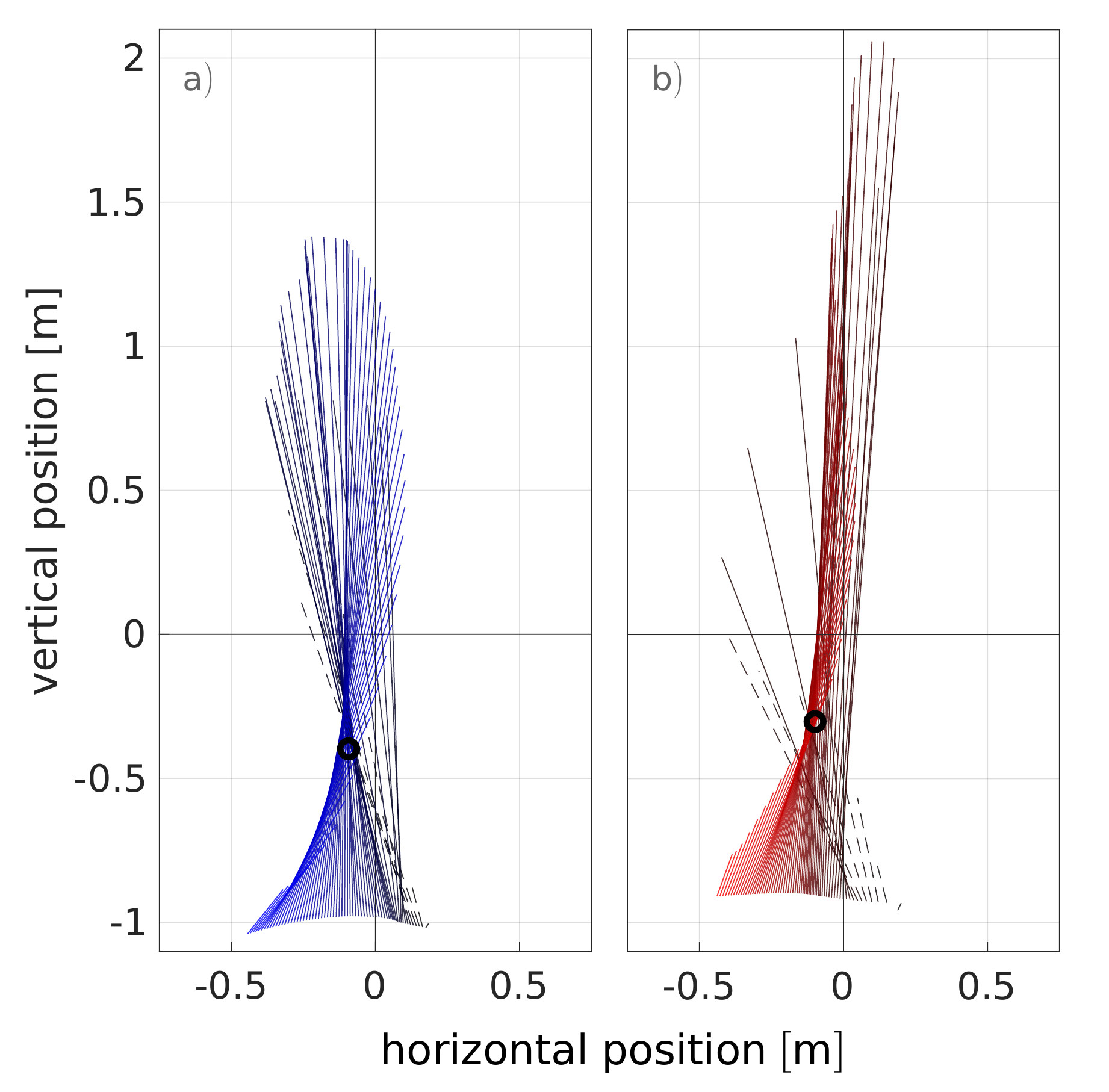}}
  {
    \caption{Examples of the ground reaction force vectors (GRFs) and the estimated virtual point (VP) for step~0 of V0 (a) and C10 (b) conditions of the human running experiments. The GRFs and VP are plotted with respect to a CoM-centered, stationary coordinate frame. Lines show the GRFs at different measurement times, originating at the CoP.
    The 90\% dataset consists only of GRF data plotted as solid lines, the 100\% dataset includes the entire stance phase GRF data.
    The black circle indicates the calculated VP for the \protect\SI{90}{\percent} dataset. a) V0: Visible level running, black to blue, b) C10: Running with a camouflaged drop of \SI{-10}{\centi\meter}, black to red. For each condition, the trial with the spread around the VP nearest to the 50\textsuperscript{th} percentile of all subjects was chosen.}
    \label{fig:VP_Example}
  }
  \end{floatrow}
  \vspace{\floatsep}%
  \begin{floatrow}
  \ffigbox[\Xhsize]
  {\includegraphics[width=0.97\linewidth]{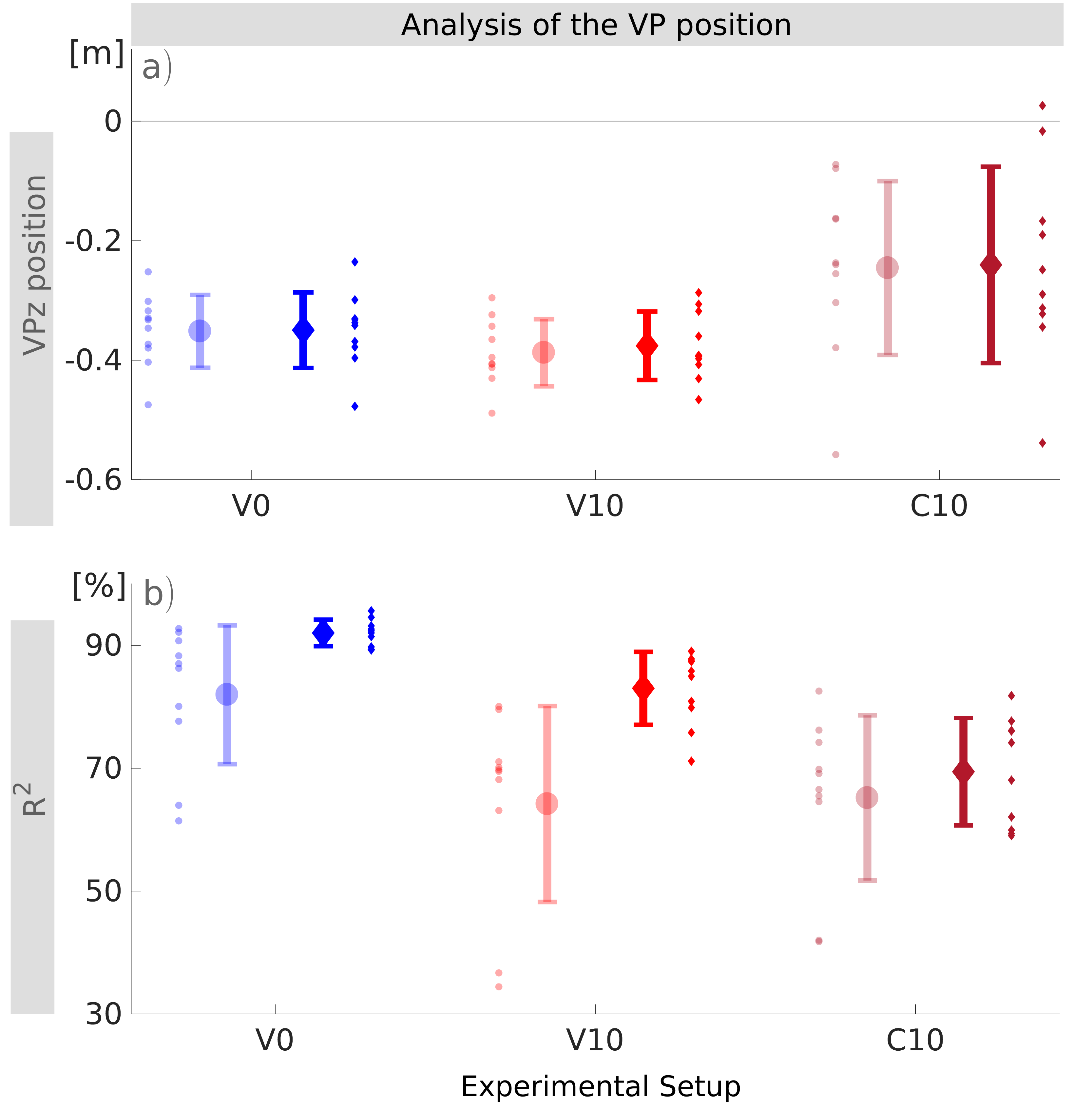}}{
    \caption{Mean$\pm$s.d. of the vertical virtual point position VPz (a), and $R^{2}$ values (b) between subjects (N=10) for each ground condition (V0, V10 and C10) for step 0.
    a) Each small dot is the median over all trials of one condition for one subject.
    b) $R^{2}$ represents the ratio of the angle between measured and ideal forces and their variance. Each small dot represents one subject.
    Transparent circle: \SI{100}{\percent} dataset, non-transparent diamond: \SI{90}{\percent} dataset.}
    \label{fig:VP_mean}
  }
  \end{floatrow}
  \vspace{-2\floatsep}%
\end{figure}

The results and statistical values of the experiments are listed in \creft{tab:VP_statistical}{} and illustrated in \crefs{fig:VP_Example}{}{ to }{fig:grf_step2}{}, and connected with simulation results, in \crefs{fig:ExpVsModel_States}{}{ to }{fig:ExpVsModel_Epot}{}.
Additionally, significant mean differences will be highlighted in the following.

In \creff{fig:VP_Example}{}, exemplary illustrations of the VP for single trials (V0 and C10) of different subjects at step~0 are shown.
Here, the GRF vectors are plotted in a CoM-centered coordinate frame were the vertical axis is parallel to gravity.
The VP is calculated as the point which minimizes the sum of squared perpendicular distances to the GRFs for each measurement time point.
To avoid biases caused by the impact peak, the VP was additionally calculated for only \SI{90}{\percent} of the dataset.
That means that the GRFs of the first \SI{10}{\percent} of the stance phase (dashed lines) were neglected in this VP calculation (\creff{fig:VP_Example}).
Hence, the VP was computed for \SI{90}{\percent} and \SI{100}{\percent} dataset and the results for both VP are given in this section.

The VP was in step~-1 (pre-perturbed) and step~0 (perturbed) below the CoM (p<0.001) and between $\minus38.8$ $\pm$ \SI{5.6}{\centi\meter} and $\minus24.0$ $\pm$ \SI{16.4}{\centi\meter} (\creff{fig:VP_mean}{a}).
For step -1, there were no differences between the ground conditions in the vertical VP position VPz (\SI{-31.0}{\centi\meter}) and the $R^{2}$ (\SI{88.7}{\percent}; \creft{tab:VP_statistical}{}).
However, the horizontal VP position VPx was \SI{5.5}{\centi\meter} (V10) and \SI{5.7}{\centi\meter} (C10) more posterior in the drop conditions than in the level condition (p<0.001).
At step 0, VPx was \SI{4.4}{\centi\meter} more posterior in C10 compared to V0 (p<0.028) and for the 100\% dataset \SI{0.8}{\centi\meter} more posterior in V10 than in V0 (p=0.038; \creft{tab:VP_statistical}{}).
There were only differences in VPz for the 100\% dataset, it was \SI{3.6}{\centi\meter} lower in V10 compared to V0 (p=0.029).
$R^{2}$ has the largest value for V0 (92.0 $\pm$ 2.1\%; 90\% dataset) and the smallest one for C10 (64.1 $\pm$ 8.7\%; 100\% dataset, \creff{fig:VP_mean}b). 

\begin{table*}[!tbh]
\caption{\label{tab:VP_statistical}Statistical analysis of VP, $R^{2}$, impulse and gait properties.
V0, visible level running; V10 visible drop of \SI{-10}{\centi\meter}; C10, camouflaged drop of \SI{-10}{\centi\meter};
VP, horizontal (x) and vertical (z) positions of the virtual point relative to the center of mass for the 90\% and the 100\% dataset;
$R^{2}$, coefficient of determination of the angles between measured ground reaction forces and forces through center of pressure and VP;
$\vec{p}_\mathrm{brake}$, braking impulse and $\vec{p}_\mathrm{prop}$, propulsion impulse in the x- and z-direction.
Data are means $\pm$ s.d. across all included subjects (N=10; exception: duty factor is only calculated for 9 subjects) for step -1 (pre-perturbed contact) and step 0 (perturbed contact).
Post hoc analysis with \v{S}id\'{a}k correction revealed significant differences
between ground conditions: differences from V0 and V10 are indicated with 'a' and 'b', respectively $(P<0.05)$.}
\centering{}%
\begin{tabular}{llr@{\extracolsep{0pt}$\pm$}lr@{\extracolsep{0pt}$\pm$}lr@{\extracolsep{0pt}$\pm$}lcc}
  &  & \multicolumn{2}{c}{\textbf{V0}} & \multicolumn{2}{c}{\textbf{V10}} & \multicolumn{2}{c}{\textbf{C10}} & p-value & F-Value/$\eta^2$\tabularnewline
  \hline
  \multirow{10}{*}{\begin{turn}{90}
  \textbf{Step -1}
  \end{turn}}
  & \cellcolor{gray!25} \textbf{VP variables} \tabularnewline
  &$\mathrm{VPx_{100\%}}$ [\SI{\centi\meter}]  & -2.9 & 2.9  & -8.5 & 3.5$^{a}$  & -8.6 & 3.1$^{a}$  & \textbf{0.000}  & 224.38/0.01\tabularnewline
  &$\mathrm{VPx_{90\%}}$ [\SI{\centi\meter}]  & -3.4 & 2.8  & -8.7 & 3.4$^{a}$  & -9.1 & 3.2$^{a}$  & \textbf{0.000}  & 146.41/0.01\tabularnewline
  &$\mathrm{VPz_{100\%}}$ [\SI{\centi\meter}]  & -31.5 & 4.9  & -31.3 & 5.0  & -31.7 & 6.6  & 0.965  & 0.04/0.00\tabularnewline
  &$\mathrm{VPz_{90\%}}$ [\SI{\centi\meter}]  & -30.8 & 5.8  & -30.7 & 5.2  & -31.5 & 6.5  & 0.997  & 0.23/0.00\tabularnewline
  &$R_\mathrm{100\%}^{2}$  [\SI{\percent}\!]  & 76.0 & 14.6  & 79.0 & 12.1  & 77.3 & 13.2  & 0.424  & 0.90/0.00\tabularnewline
  &$R_\mathrm{90\%}^{2}$  [\SI{\percent}\!]  & 88.1 & 3.4  & 89.4 & 3.4  & 88.5 & 3.1  & 0.411  & 1.45/0.00\tabularnewline
  &\cellcolor{gray!25} \textbf{Impulse} \tabularnewline
  &$\vec{p}_\mathrm{brake,x}$  & -0.05 & 0.02  & -0.05 & 0.02  & -0.04 & 0.02  & 0.162  & 2.02/0.00\tabularnewline
  &$\vec{p}_\mathrm{brake,z}$  & 0.53 & 0.11  & 0.47 & 0.10  & 0.49 & 0.06  & 0.051  & 3.53/0.01\tabularnewline
  &$\vec{p}_\mathrm{prop,x}$  & 0.11 & 0.01  & 0.12 & 0.02  & 0.11 & 0.01  & 0.078  & 2.94/0.00\tabularnewline
  &$\vec{p}_\mathrm{prop,z}$  & 0.56 & 0.02  & 0.57 & 0.04  & 0.55 & 0.04  & 0.421  & 0.91/0.00\tabularnewline
  \hline
  \multirow{16}{*}{\begin{turn}{90}
  \textbf{Step 0}
  \end{turn}}
  &\cellcolor{gray!25} \textbf{VP variables} \tabularnewline
  &$\mathrm{VPx_{100\%}}$ [\SI{\centi\meter}]  & -2.8 & 4.5  & -4.0 & 4.6$^{a}$  & -7.1 & 5.1$^{a}$  & \textbf{0.014}  & 7.95/0.01\tabularnewline
  &$\mathrm{VPx_{90\%}}$ [\SI{\centi\meter}]  & -2.6 & 4.6  & -4.3 & 4.7  & -7.0 & 5.0$^{a}$  & \textbf{0.018}  & 7.17/0.01\tabularnewline
  &$\mathrm{VPz_{100\%}}$ [\SI{\centi\meter}]  & -35.2 & 6.1  & -38.8 & 5.6$^{a}$  & -24.6 & 14.5  & \textbf{0.047}  & 5.17/0.10\tabularnewline
  &$\mathrm{VPz_{90\%}}$ [\SI{\centi\meter}]  & -35.0 & 6.3  & -37.6 & 5.7  & -24.0 & 16.4  & 0.074  & 4.04/0.10\tabularnewline
  &$R_\mathrm{100\%}^{2}$  [\SI{\percent}\!] & 81.9 & 11.3  & 64.1 & 15.9$^{a}$  & 65.1 & 13.4  & \textbf{0.021}  & 6.87/0.17\tabularnewline
  &$R_\mathrm{90\%}^{2}$  [\SI{\percent}\!]  & 92.0 & 2.1  & 83.0 & 5.9$^{a}$  & 69.4 & 8.7$^{a,b}$  & \textbf{0.000}  & 70.13/0.13\tabularnewline
  &\cellcolor{gray!25} \textbf{Impulse} \tabularnewline
  &$\vec{p}_\mathrm{brake,x}$  & -0.10 & 0.02  & -0.11 & 0.03  & -0.04 & 0.02$^{a,b}$  & \textbf{0.000}  & 40.27/0.01\tabularnewline
  &$\vec{p}_\mathrm{brake,z}$  & 0.69 & 0.08  & 0.83 & 0.12$^{a}$  & 0.63 & 0.12$^{b}$  & \textbf{0.000}  & 20.92/0.10\tabularnewline
  &$\vec{p}_\mathrm{prop,x}$  & 0.09 & 0.02  & 0.09 & 0.01  & 0.06 & 0.01$^{a,b}$  & \textbf{0.000}  & 14.26/0.00\tabularnewline
  &$\vec{p}_\mathrm{prop,z}$  & 0.46 & 0.08  & 0.48 & 0.05  & 0.45 & 0.06  & 0.309  & 1.19/0.01\tabularnewline
  &\cellcolor{gray!25} \textbf{Gait properties} \tabularnewline
  &velocity [\SI{}{\meter\per\second}]  & 4.9 & 0.5  & 4.9 & 0.5  & 5.1 & 0.4  & 0.148  & 2.13/0.11\tabularnewline
  &stance time [\SI{\second}]  & 0.18 & 0.02  & 0.17 & 0.02$^{a}$   & 0.14 & 0.01$^{a,b}$   & \textbf{0.000}  & 62.67/0.00\tabularnewline
  &duty factor [\SI{\percent}\!] & 26.7 & 2.0  & 24.8 & 1.6$^{a}$  & 22.4 & 1.5$^{a,b}$  & \textbf{0.008}  & 37.20/0.01 \tabularnewline
\end{tabular}
\end{table*}

\begin{figure}[tb!]
\centering
\includegraphics[width=1\linewidth]{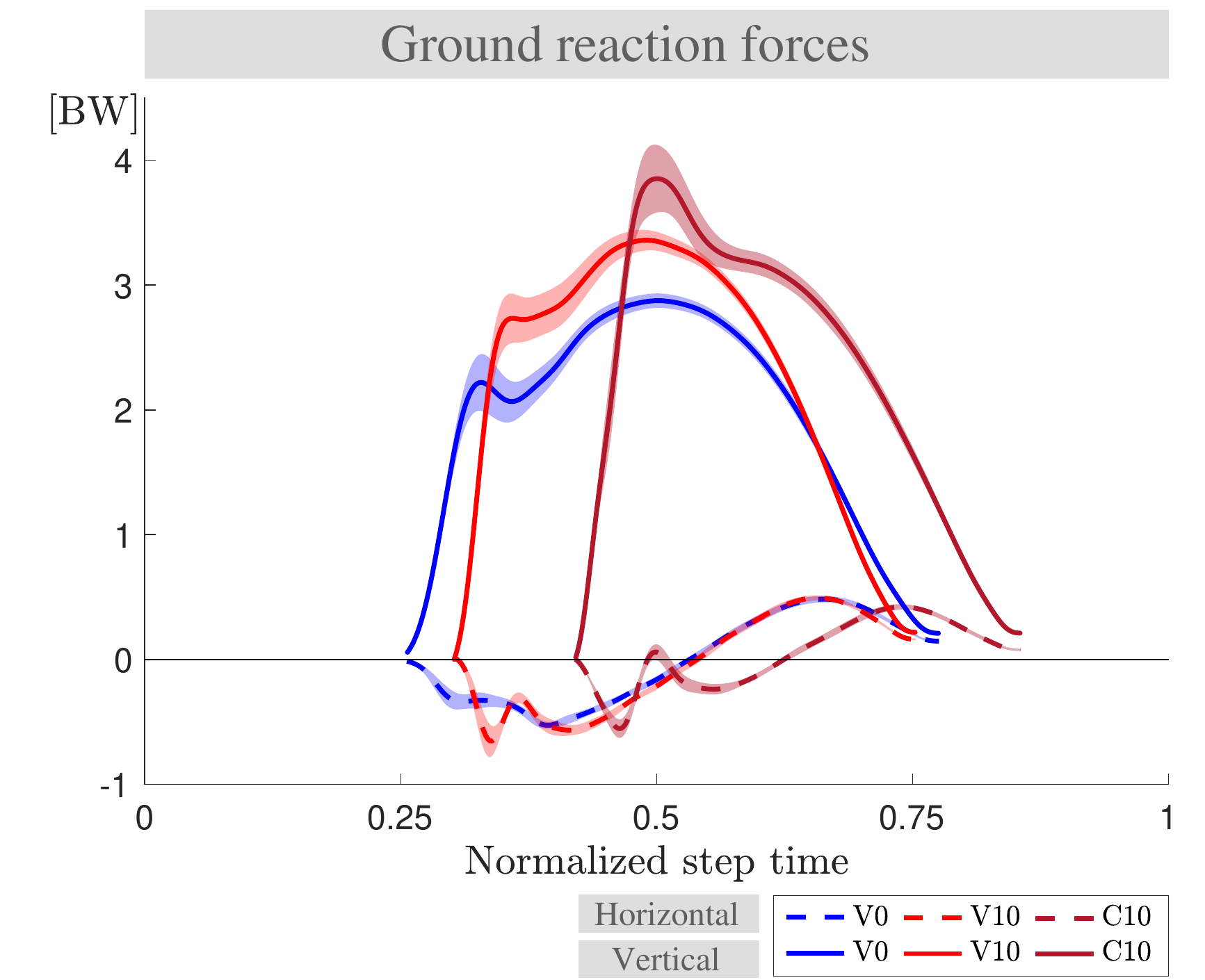}
\caption{The ground reaction forces (GRFs) of step~0 for human running experiments V0 (blue), V10 (red), and C10 (brown). The GRF are normalized to body weight of the subjects (N=10) .
The mean values of the vertical and horizontal GRF are plotted with solid and dashed lines, respectively. The $\protect\pm$ standard error is shown with the shaded area.
For the C10 condition, the vertical GRF peak coincides with the peak caused by the impact peak forces.
The duty factor of the V0 condition is \protect\SI{26.7 \pm 2.0}{\percent}, whereas it is \protect\SI{24.8 \pm 1.6}{\percent} for the V10 condition and \protect\SI{22.4 \pm 1.5}{\percent} for the C10 condition.
%
}\label{fig:grf_step2}
\end{figure}

There were no significant differences between the ground conditions in the impulses of step -1 (\creft{tab:VP_statistical}{}).
For step 0, \creff{fig:grf_step2}~suggests that the vertical GRFs are higher in the step conditions compared to V0, especially for the braking phase.
The vertical braking impulse was higher in V10 than in V0 (p=0.008) and in C10 (p<0.001).  
We observe \SI{2.9}{\bw} peak vertical GRFs in V0, which yield to a vertical braking impulse of 0.69.
In V10, the peak vertical GRFs were at \SI{3.4}{\bw} with a braking impulse of 0.83.
In C10, the peak was the highest with \SI{3.9}{\bw}, but here, the peak is overlapping with the impact peak and therefore not comparable with that of the visible ground conditions (\creff{fig:grf_step2}).
Because of the shorter stance time in C10 (\creft{tab:VP_statistical}{}), the braking impulse of 0.63 does not differ from the value of V0 despite the high impact peak.
The vertical propulsion impulse of step 0 does not differ significantly between the ground conditions.
The amounts of the horizontal braking and propulsion impulses were lower in C10 than in the visible conditions (p$\leq$0.004).
The sum of the horizontal braking and propulsion impulses of step 0 is in all ground conditions around zero.
That means that there is no forward acceleration or deceleration. 

%
The vertical CoM position relative to the CoP at the touch-down of step 0 is \SI{3.5}{\centi\meter} higher in the drop conditions compared to V0 (p<0.001) with \SI{104.9 \pm 5.2}{\centi\meter} and \SI{1}{\centi\meter} higher in C10 than in V10 (p=0.019).

The forward running velocity measured at step~0 does not vary between the experiments V0, V10 and C10, and is within the range of \SI{5.0 \pm 0.5}{\meter\per\second}.
Despite the constant velocity, the stance time and the duty factor of step~0 show a variation for between these experiments.
The stance time gets shorter (p=0.029) and the duty factor lower (p<0.001) when running down the visible drop and even shorter and lower when the drop is camouflaged (p<0.006).

%% file: sec/result_sim.tex
In this section, we present our simulation results and our analysis on how VP reacts to step-down perturbations.
%
The simulation gaits are generated for \SI{5}{\meter\per\second} running with a VP target \SI{\minus 30}{\centi\meter} below the CoM (\VPbl), which correspond to the estimated values of our experiments in \cref{subsec:experimentalresults}.

\begin{figure*}[!bt]
\centering
\begin{annotatedFigure}
	{\includegraphics[width=1\linewidth]{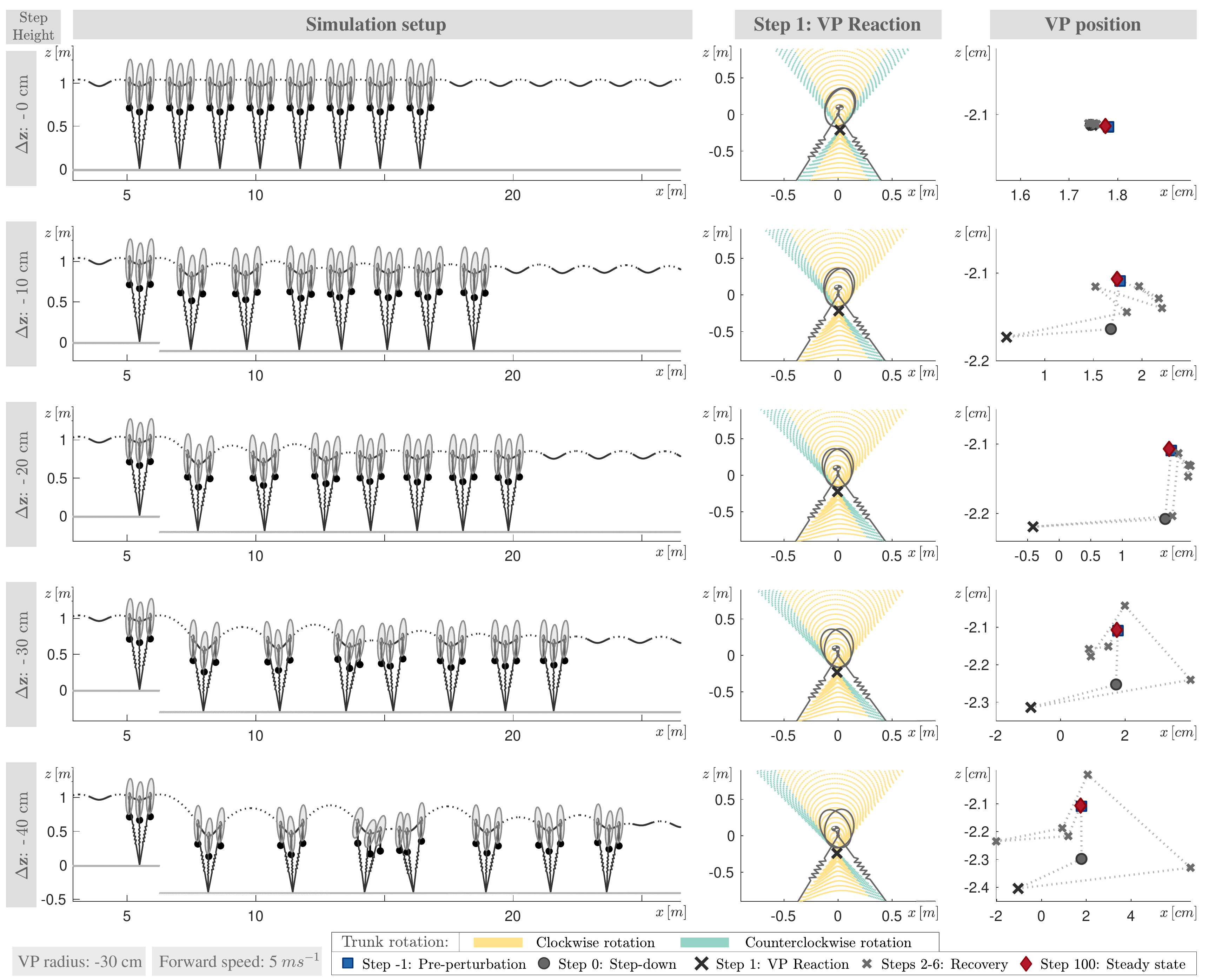}}
	\sublabel{$a_{0}$)}{0.08,0.945}{color_gray}{0.7}  \sublabel{$b_{0}$)}{0.55,0.945}{color_gray}{0.7} \sublabel{$c_{0}$)}{0.77,0.945}{color_gray}{0.7}
	\sublabel{$a_{1}$)}{0.08,0.76}{color_gray}{0.7}    \sublabel{$b_{1}$)}{0.55,0.76}{color_gray}{0.7}   \sublabel{$c_{1}$)}{0.77,0.76}{color_gray}{0.7}
	\sublabel{$a_{2}$)}{0.08,0.58}{color_gray}{0.7}    \sublabel{$b_{2}$)}{0.55,0.58}{color_gray}{0.7}  \sublabel{$c_{2}$)}{0.77,0.58}{color_gray}{0.7}
	\sublabel{$a_{3}$)}{0.08,0.395}{color_gray}{0.7} \sublabel{$b_{3}$)}{0.55,0.395}{color_gray}{0.7} \sublabel{$c_{3}$)}{0.77,0.395}{color_gray}{0.7}
	\sublabel{$a_{4}$)}{0.08,0.21}{color_gray}{0.7}    \sublabel{$b_{4}$)}{0.55,0.21}{color_gray}{0.7}    \sublabel{$c_{4}$)}{0.77,0.21}{color_gray}{0.7}
\end{annotatedFigure}
\captionof{figure}{ The analysis begins with a base gait $a_{0}$ based on the human running experiment V0, which has a VP target of \SI{-30}{\centi\meter} with a forward speed of \SI{5}{\meter\per\second}. This base gait is then subjected to step-down perturbations of $\Delta z \myeq$[\SI{ \protect\minus 10, \protect\minus 20,  \protect\minus 30, \protect\minus 40}{\centi\meter}] at step~0. The \protect\SI{\protect \minus 10}{\centi\meter} perturbation corresponds to V10\,-\,C10 of the human running experiments. The model state at touch-down, mid-stance and take-off instances of steps~-1 to 6 are drawn in $a_{0}  \protect \minus a_{4}$ to display the changes in the trunk angle.
At the perturbation step, the VP position shifts downward with respect to a hip centered stationary coordinate frame ({\protect \markerCircle[color_darkgray]} in $c_{1} \protect \minus c_{4}$).
\VPbl counteracts to the perturbation at step~1 with a left shift, which depletes the energy added by the stepping down ({\protect \raisebox{-0.6 pt}{\markerCross[color_darkgray][1.3]}} in $c_{1} \protect \minus c_{4}$). The GRF vectors of step~1 causes a forward trunk lean of 5 to \protect\SI{10}{\degree}, which is shown in $b_{1} \protect \minus b_{4}$. In the following steps, VP position is regulated to achieve the energy balance ({\protect \markerCross[color_gray][1]}), and gaits ultimately reach to the equilibrium state{\protect\footnotemark}\,({\protect \raisebox{-0.3 pt}{\markerDiamond[color_red][0]}} markers in figure $c_{1} \protect \minus c_{4}$).
%
}\label{fig:SimulationResults}
\end{figure*}

The temporal properties of the {\color{color_gray}{base gait}} for the level running are given in \creft{tab:GaitProperties}{}, where the duty factor is calculated as \SI{26.2}{\percent} with a stance phase duration of \SI{0.16}{\second}. The CoM trajectory of the base gait is shown in \creff{fig:SimulationResults}{$a_{0}$} and its respective GRF vectors are plotted with respect to a hip centered stationary coordinate frame in \creff{fig:SimulationResults}{$b_{0}$}.

\begin{table}[!bt]
\centering
\captionsetup{justification=justified}
\caption{Gait properties of the simulated trajectories. In the presence of step-down perturbations, the \VPbl method is able to bring the system back to its initial equilibrium state. Therefore, the gait properties are the same for the even ground and perturbed terrain, after reaching to the steady state condition}.
\label{tab:GaitProperties}
\begin{adjustbox}{width=0.98\columnwidth}
\begin{tabular}{@{} l| c c|| c c c  @{}}
\multicolumn{1}{l}{Property} & \multicolumn{1}{c}{Unit} &  \multicolumn{1}{c||}{Value} &   \multicolumn{1}{c}{Property} & \multicolumn{1}{c}{Unit} & \multicolumn{1}{c}{Value}  \\
\hline
\hspace{1mm} Duty factor & [$\%$]   &  26.2  &   VP angle & [\si{\degree}] &  -180   \\
\hspace{1mm} Stance time  &[s] &  0.16 &    Trunk angular excursion & [\si{\degree}] &  4.45   \\
\hspace{1mm} Forward speed & [\si{\meter\per\second}] &  5 &    Leg angle at touch-down & [\si{\degree}] &  66   \\
\end{tabular}
\end{adjustbox}
\end{table}

The base gait is subjected to step-down perturbations of  $\Delta z \myeq$[\SI{ \protect\minus 10, \protect\minus 20,  \protect\minus 30, \protect\minus 40}{\centi\meter}] at step~0.
%
The VP controller updates on step-to-step basis, therefore it is informed about the deviation from the base gait at the beginning of step~1.
%
In step~1, the VP controller shifts the \VPbl to the left as seen with {\protect \markerCircle[color_darkgray]} marker in \crefsub{fig:SimulationResults}{$c_{1}$}{$c_{4}$}.
%
The leftward \VPbl shift leads to a more pronounced forward trunk motion at step~1, as can be seen in the absence of a counterclockwise rotation towards the leg take-off, i.e., the GRF vectors are not colored teal {\protect\markerRectangle[color_teal]} towards leg take-off in \crefsub{fig:SimulationResults}{$b_{1}$}{$b_{4}$}, in contrast to \creff{fig:SimulationResults}{$b_{0}$}.
%
We see that the \VPbl is able to counteract the step-down perturbations in the following steps by using only local controllers for the VP angle (\crefe{eqn:thetaVP}) and the leg angle (\crefe{eqn:thetaL}), as shown in \crefsub{fig:SimulationResults}{$a_{1}$}{$a_{4}$}.
%
As we increase the magnitude of the step-down perturbations, we decrease the coefficients $k_{\dot{x}}, k_{\dot{x}_{0}}$ in the leg angle control, so that the speed correction is slower and the postural control is prioritized (see \cref{sec:app:simTSLIPparameters}).
%
The generated gaits are able to converge to the initial equilibrium state (i.e., the initial energy level) within 15 steps after the step-down perturbation at step~0.
%

\begin{figure}[tb!]
{\captionof{figure}{ The energy levels for the leg spring (a), leg damper (b) and hip actuator (c) for \SI{\protect\minus 10}{\centi\meter} step-down perturbation. The step-down perturbation at step~0 increases the energy of the system, which causes an increase in leg deflection and a larger fluctuation in spring energy (a, {\protect\markerLine[color_darkgray]}). The leg damper dissipates more energy and the hip actuator injects more energy than during its equilibrium condition (b-c, {\protect\markerLine[color_darkgray]}). Starting with step~1, the VP begins to react to the energy change and the hip actuator starts to remove energy from the system (c, {\protect\markerDashedLine[color_darkgray]}). In the following steps ({\protect\markerLine[color_gray]}) the hip regulates the energy until the system reaches to the initial equilibrium state ({\protect\markerLine[color_blue]}).  Extended plots for the step-down height of $\Delta z \protect\myeq$[\SI{ \protect\minus 20,  \protect\minus 30, \protect\minus 40}{\centi\meter}] can be found in \cref{sec:app:simLegHipEnergies}.
}\label{fig:Energy_dt}}
{\begin{annotatedFigure}
	{\includegraphics[width=1\linewidth]{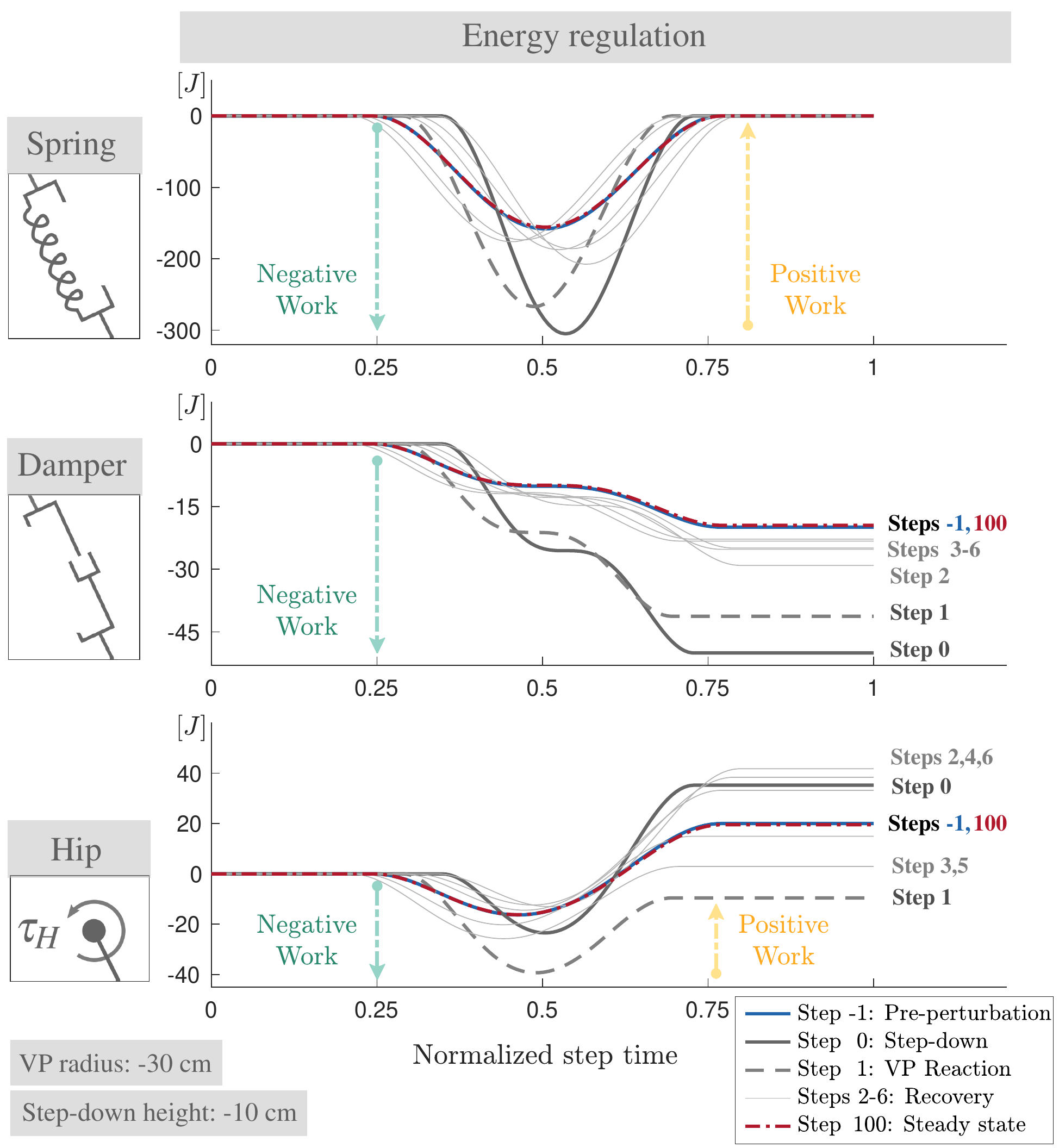}}
	\sublabel{a)}{0.15,0.92}{color_gray}{1}
	\sublabel{b)}{0.15,0.64}{color_gray}{1}
	 \sublabel{c)}{0.15,0.365}{color_gray}{1}
\end{annotatedFigure}}
\end{figure}

\footnotetext{The equilibrium state is given in \creft{tab:GaitProperties}{}. A single gait involves 100 successful steps.}

\subsubsection{Energy regulation }\label{subsubsec:simEnergy}
In order to assess the response of the VP controller, we plot the VP position with respect to a hip centered non-rotating coordinate frame that is aligned with the global vertical axis, as it can be seen in \crefsub{fig:SimulationResults}{$c_{1}$}{$c_{4}$}.
%
For a \VPbl target, a left shift in VP position indicates an increase in the negative hip work.
%

The step-down perturbation at step~0 increases the total energy of the system by the amount of potential energy introduced by the perturbation, which depends on the step-down height.
%
The position of the VP with respect to the hip shifts downward by \SI{0.5 \minus 1.9}{\centi\meter} depending on the drop height (see circle markers\,{\protect \markerCircle[color_darkgray]} in \crefsub{fig:SimulationResults}{$c_{1}$}{$c_{4}$}).
Consequently, the net hip work remains positive and its magnitude increases by $0.7 \text{ to } 1.7$ fold\footnote{For quantities A and B, the fold change is given as $\mathrm{(B \minus A) / A}$.} (see solid lines\,{\protect\markerLine[color_darkgray]} in \crefs{fig:Energy_dt}{c}{~and~}{fig:Energy_dt_Dz}{c}).
The leg deflection increases by $0.95 \text{ to } 3$ fold, whose value is linearly proportional to the leg spring energy as $E_{SP} \myeq \frac{1}{2} \, k\, \Delta l_{L}^{2}$ (see solid lines\,{\protect\markerLine[color_darkgray]} in \crefs{fig:Energy_dt}{a}{~and~}{fig:Energy_dt_Dz}{a}).
%
The leg damper dissipates $1.5 \text{ to } 6$ fold more energy compared to its equilibrium condition (see solid lines\,{\protect\markerLine[color_darkgray]} in \crefs{fig:Energy_dt}{b}{~and~}{fig:Energy_dt_Dz}{b}).

The reactive response of the VP starts at step~1, where the target VP is shifted to left by \SI{1.2 \minus 2.8}{\centi\meter} and down by \SI{0.6 \minus 2.9}{\centi\meter} depending on the drop height (see cross markers\,{\protect \raisebox{-0.6 pt}{\markerCross[color_darkgray][1.3]}} in \creff{fig:SimulationResults}{c}).
The left shift in VP causes a $1.4 \text{ to } 3.8$ fold increase in the negative hip work, and the \emph{net} hip work becomes negative (see dashed lines\,{\protect\markerDashedLine[color_darkgray]} in \crefs{fig:Energy_dt}{c}{~and~}{fig:Energy_dt_Dz}{c}). In other words, the hip actuator starts to remove energy from the system.
As a result, the trunk leans more forward during the stance phase (see yellow colored GRF vectors {\protect\markerRectangle[color_yellow]} in \creff{fig:SimulationResults}{b}).
The leg deflects $0.7 \text{ to } 2.3$ fold larger than its equilibrium value, and the leg damper removes between $1 \text{ and } 4.1$ fold more energy. However, the increase in leg deflection and damper energy in step~1 are lower in magnitude compared to the increase in step~0.
In step~1, we see the \VPbl's capability to remove the energy introduced by the step-down perturbation.

In the steps following step~1, the target VP position is continued to be adjusted with respect to the changes in the trunk angle at apices, as expressed in \crefe{eqn:thetaVP} and shown with {\protect \markerCross[color_gray][1]} markers in \creff{fig:SimulationResults}{c}. The VP position gradually returns to its initial value, and the gait ultimately converges to its initial equilibrium, see coinciding markers {\protect \raisebox{-0.3 pt}{\markerDiamond[color_red][0]},\,{\protect \markerSquare[color_blue][0]} in \creff{fig:SimulationResults}{c}}. During this transition, the energy interplay between the hip and leg successfully removes the energy added to the system, as shown in \crefsub{fig:Energy_dt}{$b$}{$c$} and in \crefsub{fig:Energy_dt_Dz}{b}{c} for larger step-down perturbation magnitudes.

\subsubsection{GRF analysis}\label{subsubsec:simEnergy}
\begin{figure}[tb!]
{\captionof{figure}{Numerical simulation results: The ground reaction forces (a,c) and the corresponding net impulses (b,d) for -\SI{10}{\centi\meter} step-down perturbation. The GRFs are normalized to body weights (\si{\bw}), whereas the impulses are normalized to their $\mathsmaller{\mathrm{BW} \mathsmaller{\mathop{\sqrt{\sfrac{l}{g}}}}}$ values. The effect of the \VPbl control can be seen in the horizontal GRF and impulse. \VPbl alters the net horizontal impulse, and causes either net horizontal acceleration or deceleration after the step-down perturbation. Consequently, the excess energy introduced by the perturbation is removed from the system. The vertical GRF and impulse increase with the perturbation and decrease gradually to its equilibrium value approximately within 15 steps. Extended plots for the step-down height of $\Delta z \protect\myeq$[\SI{ \protect\minus 20,  \protect\minus 30, \protect\minus 40}{\centi\meter}] can be found in \cref{sec:app:simGRF}.
}\label{fig:GRF}}
{\begin{annotatedFigure}
	{\includegraphics[width=1\linewidth]{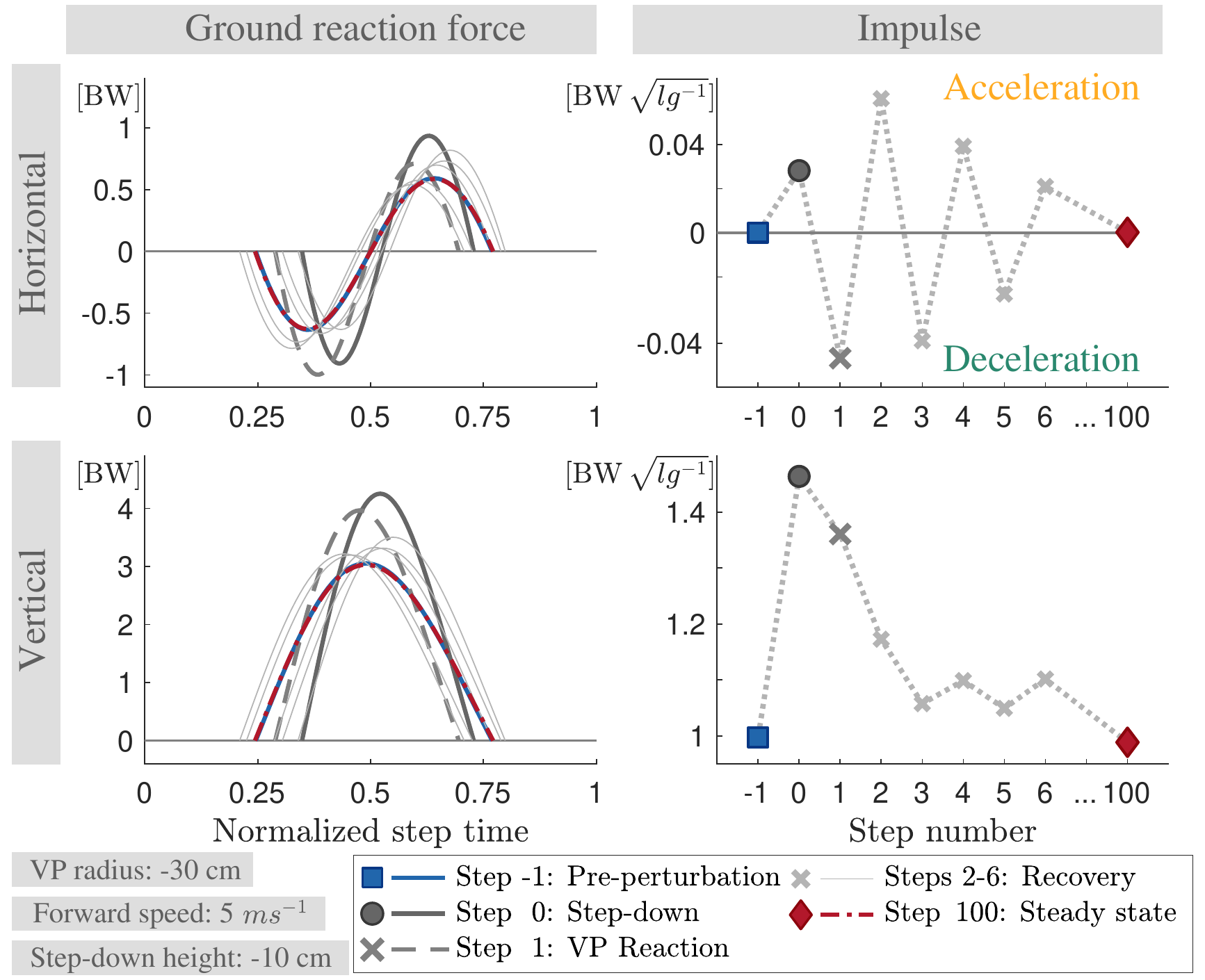}}
	\sublabel{a)}{0.15,0.92}{color_gray}{1}
	\sublabel{c)}{0.15,0.53}{color_gray}{1}
	 \sublabel{b)}{0.62,0.92}{color_gray}{1}
	  \sublabel{d)}{0.62,0.53}{color_gray}{1}
\end{annotatedFigure}}
\end{figure}

The energy increment due to the step-down perturbation and the energy regulation of the \VPbl control scheme can also be seen in the GRF and impulse profiles.

The  peak vertical GRF magnitude of the equilibrium state is \SI{3}{\bw}. It increases to \SI{4.2 \minus 6.1}{\bw} at step~0 with the step-down (\crefs{fig:GRF}{c}{~and~}{fig:GRFy}{a}). The peak magnitude decreases gradually to its initial value in the following steps, indicating that the VP is able to bring the system back to its equilibrium.
%
In a similar manner, the normalized vertical impulse increases from $1$ to $1.4 \minus 2.2$ at step~0 ({\protect \markerCircle[color_darkgray]}) and decreases to $1$ in approximately 15 steps.

The  peak horizontal GRF magnitude of the equilibrium state amounts to \SI{0.6}{\bw}.  It increases to \SI{0.9 \minus 1.4}{\bw} at step~0 (\crefs{fig:GRF}{a}{~and~}{fig:GRFx}{a}). The sine shape of the horizontal GRF and its peak magnitude depend on the change in VP position. Therefore, the horizontal GRF impulse provides more information.
%
The net horizontal GRF impulse is zero at the equilibrium state (see {\protect \markerSquare[color_blue][0]} in \crefs{fig:GRF}{b}{~and~}{fig:GRFx}{b}). It becomes positive at the step-down perturbation ({\protect \markerCircle[color_darkgray]}), leading to a net horizontal acceleration of the CoM. In step~1, the \VPbl is adjusted with respect to the change in the state and causes the impulse to decelerate the body ({\protect \raisebox{-0.6 pt}{\markerCross[color_darkgray][1.5]}}).
%
In the following steps, the VP adjustment yields successive net accelerations and decelerations ({\protect \markerCross[color_gray][1]}) until the system returns to its equilibrium state ({\protect \raisebox{-0.3 pt}{\markerDiamond[color_red][0]}}).

%% file: sec/discussion.tex

In this study, we performed an experimental and numerical analysis regarding the force direction patterns during human level running, and running onto a visible or camouflaged step-down.
Our experimental results show that humans tend to generate a VP below the CoM (\VPb) for all terrain conditions.
Our simulations support this experimental observations, and show that the \VPb as a controller can cope with step-down perturbations up to 0.4 times the leg length.
%
In this section, we will address the VP location in connection with the gait type,
and will discuss how our experimental results compare to our simulation results for the running gait.

\subsection{VP quality and location in human gait}\label{subsec:VPingait}
In the first part, we discuss the validity of a virtual \emph{point} estimated from the GRF measurements of the human running.
We only consider step~0 of the \SI{90}{\percent} dataset, since the \SI{100}{\percent} dataset is biased by the additional effects of the impact forces and has low $R^2$ values \cite{blickhan2015positioning}. In the second part, we discuss how the VP position is correlated to the gait type.

To determine the quality of the virtual point estimation, we used the coefficient of determination $R^2$. In our experiments, the $R^2$ values for level running are high, where $R^2\!\approx\,$\SI{92}{\percent} (see V0 in \creff{fig:VP_mean}{b}).
%
The values of the $R^2$ get significantly lower for the visible drop condition, where $R^2\!\approx\,$\SI{83}{\percent} (see V10 in \creff{fig:VP_mean}{b}).
On the other hand, the $R^2$ of the camouflaged drop conditions are even lower than for the visible drop conditions, where $R^2\!\approx\,$\SI{69}{\percent} (see C10 in \creff{fig:VP_mean}{b}). An $R^2$ value of $\approx\,$\SI{70}{\percent} is regarded as "reasonably well" in the literature \cite[p.475]{herr2008angular}.
Based on the high $R^2$ values, we conclude that the measured GRFs intersect near a \emph{point} for the visible and camouflaged terrain conditions.
We can also confirm that this point is below the CoM (\VPb), as the mean value of the estimated points is \SI{\minus 32.2}{\centi\meter} and is significantly below the CoM.
%

We find a difference in the estimated VP position between the human walking and our recorded data of human running.
%
The literature reports a VP above the CoM (\VPa) for human walking gait \cite{gruben2012force,maus2010upright,muller2017force,vielemeyer2019ground}, some of which report a \VPa in human running as well \cite{Maus_1982,blickhan2015positioning}.
In contrast, our experiments show a \VPb for human level running at \SI{5}{\meter\per\second} and running over a visible or camouflaged step-down perturbation.
%
Our experimental setup and methodology are identical to \cite{vielemeyer2019ground}, which reports results from human walking. Thus, we can directly compare the $R^2$ values for both walking and running.
%
The $R^2$ value of the level running is 6~percentage points lower than the $R^2$ reported in \cite{vielemeyer2019ground} for level walking.
%
The $R^2$ value for V10 running is 15~percentage points lower than V10 walking, whereas the $R^2$ for C10 running is up to 25~percentage points lower compared to C10 walking.
%
In sum, we report that the spread of the $R^2$ is generally higher in human running at \SI{5}{\meter\per\second}, compared to human walking.

\subsection{Experiments vs. model}\label{subsec:modelVSexp}
\begin{figure}[t!]
{\captionof{figure}{The CoM height (a), horizontal CoM velocity (b) and trunk angle (c) for the step~0 of the experiments V0, V10 and C10 are shown on left\protect\footnotemark, and the steps -1, 0 and 1 of the simulation are shown on right column. The TSLIP model is able to predict the CoM height and forward speed. Its prediction capability is reduced for the trunk motion, as the flight phase involves ballistic motion and the trunk angular velocity is constrained to be constant.
}\label{fig:ExpVsModel_States}}
{\begin{annotatedFigure}
	{\includegraphics[width=1\linewidth]{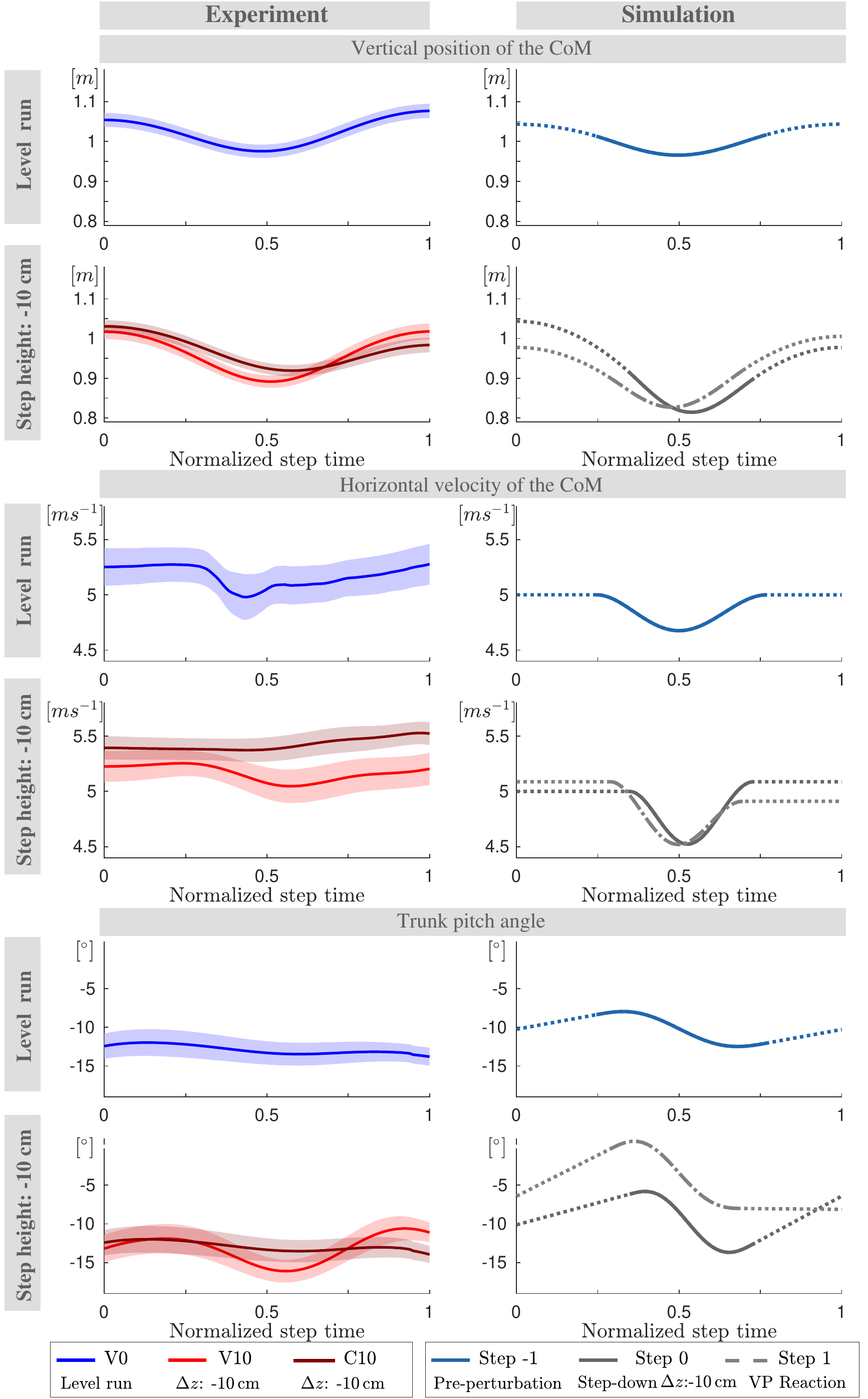}}
	\sublabel{$a_{1}$)}{0.16,0.945}{color_gray}{0.8}  \sublabel{$a_{2}$)}{0.63,0.945}{color_gray}{0.8}
	\sublabel{$a_{3}$)}{0.16,0.805}{color_gray}{0.8}  \sublabel{$a_{4}$)}{0.63,0.805}{color_gray}{0.8}
	\sublabel{$b_{1}$)}{0.16,0.635}{color_gray}{0.8}  \sublabel{$b_{2}$)}{0.63,0.635}{color_gray}{0.8}
	\sublabel{$b_{3}$)}{0.16,0.495}{color_gray}{0.8}  \sublabel{$b_{4}$)}{0.63,0.495}{color_gray}{0.8}
	\sublabel{$c_{1}$)}{0.16,0.32}{color_gray}{0.8}  \sublabel{$c_{2}$)}{0.63,0.32}{color_gray}{0.8}
	\sublabel{$c_{3}$)}{0.16,0.185}{color_gray}{0.8}  \sublabel{$c_{4}$)}{0.63,0.185}{color_gray}{0.8}
\end{annotatedFigure}}
\end{figure}

In this section, we discuss how well the TSLIP simulation model predicts the CoM dynamics, trunk angle trajectories, GRFs and energetics of human running.
A direct comparison between the human experiments and simulations is possible for the level running. The V0 condition of the human experiments corresponds to step~-1 of the simulations (also to the base gait).
Overall, we observe a good match between experiments and simulations for the level running (see \crefs{fig:ExpVsModel_States}{}{~to~}{fig:ExpVsModel_Epot}).
On the other hand, a direct comparison for the gaits with perturbed step is not feasible due to the reasons given in \cref{subsec:Limitations} in detail. Here, we present perturbed gait data to show the extent of the similarities and differences between the V10 and C10 conditions of the experiments and step~0~and~1 of the simulations.

Concerning the CoM dynamics, the predicted CoM height correlates closely with the actual CoM height in level running, both of which fluctuate between \SI{1.05 \minus 1.00}{\meter}  with \SI{5}{\centi\meter} vertical displacement (\crefsub{fig:ExpVsModel_States}{$a_{1}$}{$a_{2}$}).
The vertical displacement of the CoM is larger for the perturbed step, where the CoM height alternates between \SI{1.0 \minus 0.9}{\meter} in the experiments (\creff{fig:ExpVsModel_States}{$a_{3}$}) and \SI{1.05 \minus 0.85}{\meter} in the simulations (\creff{fig:ExpVsModel_States}{$a_{4}$}).
The differences can be attributed to the visibility of the drop. Human runners visually perceiving changes in ground level and lowered their CoM by 
about 25\% of the possible drop height for the camouflaged contact \cite{ernst2019humans}.
%
The mean forward velocity at leg touch-down is \SI{5.2}{\meter\per\second} in the experiments (\creff{fig:ExpVsModel_States}{$b_{1}$}).
In the simulations, the leg angle controller adjusts the forward speed at apex to a desired value. We set the desired speed to \SI{5}{\meter\per\second} (\creff{fig:ExpVsModel_States}{$b_{2}$}), which is the mean forward velocity of the step estimated from the experiments.
%
%
For level running, both the experiments and simulations show a \SI{0.2}{\meter\per\second} decrease in forward velocity between the leg touch-down and mid-stance phases (\crefsub{fig:ExpVsModel_States}{$b_{1}$}{$b_{2}$}).
%
As for the perturbed running, human experimental running shows a drop in forward speed of \SI{4.5}{\percent} for V10, and \SI{0.1}{\percent} for the C10 condition (see \creff{fig:ExpVsModel_States}{$b_{3}$}{}).
Namely, there is no significant change in forward velocity during the stance phase for the C10 condition.
The simulation shows a drop in forward speed of \SI{9.5}{\percent} for step~0, and \SI{11.1}{\percent} in step~1 (see \creff{fig:ExpVsModel_States}{$b_{4}$}{}).

The trunk angle is the least well predicted state, since the S-shape of the simulated trunk angle is not recognizable in the human running data (see \crefsub{fig:ExpVsModel_States}{$c_{1}$}{$c_{2}$}).
One of the reasons may be the simplification of the model. The flight phase of a TSLIP model is simplified as a ballistic motion, which leads to a constant angular velocity of the trunk. The human body on the other hand is composed of multiple segments, and intra-segment interactions lead to more complex trunk motion during flight phase.
%
%
In addition, there is a large variance in the trunk angle trajectories between different subjects and trials, in particular for the C10 condition.
%
Consequently, the mean trunk angle profiles do not provide much information about the trunk motion pattern, especially for the perturbed step for C10.
%
%
Therefore, we can not clarify to what extend the VP position is utilized for regulating the trunk motion in humans.
%
%
However, a trend of trunk moving forward is visible in both simulation and experiments.
The mean trunk angular excursion at step~0 of the experiments is \SI{1.8}{\degree} for V0, \SI{5.5}{\degree} for V10, and \SI{1.9}{\degree} for the C10 condition (\crefsub{fig:ExpVsModel_States}{$c_{1}$}{$c_{3}$}).
The S-shaped pattern of the trunk motion becomes more perceivable in the experiments with a visible perturbed step (\creff{fig:ExpVsModel_States}{$c_{3}$}).
In the simulations, the trunk angular excursion is set to \SI{4.5}{\degree} for level running based on \cite{Heitcamp_2012,Schache_1999,thorstensson1984trunk}. The magnitude of the trunk rotation at the perturbation step is higher in simulations, and amounts to \SI{7.8}{\degree} at step~0 and \SI{8.6}{\degree} at step~1 (\crefsub{fig:ExpVsModel_States}{$c_{2}$}{$c_{4}$}).

\footnotetext{The mean is shown with a line and the standard error is indicated with the shaded region. The standard error equals to the standard deviation divided by the square roof of number of subjects.}

There is a good agreement between the simulation-predicted and the recorded GRFs for level running.
The peak horizontal and vertical GRFs amount to \SI{0.5}{\bw} and \SI{3}{\bw} respectively, in both experiments and simulations (see \crefsss{fig:grf_step2}{,}{\,}{fig:GRF}{a,\,}{d,\,}{and\,}{fig:GRFexpsim}{}).
As for the step-down perturbation, the simulation model is able to predict the peak vertical GRF, but the prediction becomes less accurate for the peak horizontal GRF.
The peak vertical GRF of the \SI{\minus 10}{\centi\meter} step-down perturbation case is \SI{3.5}{\bw} for the V10 condition and \SI{4}{\bw} for the C10 condition, whereas it is \SI{4}{\bw} for the simulation.
In the C10 condition, the vertical GRF peak occurs at the foot impact and its peak is shifted in time, to the left.
%
%
The numerical simulation leads to over-simplified horizontal GRF profiles, in the step-down condition. The human experiments show an impact peak.
The experiments have a peak horizontal GRF magnitude of \SI{0.5}{\bw}, which remains the same for all perturbation conditions. In contrast, the peak horizontal GRF increases up to \SI{1}{\bw} in simulations.

In level running the GRF impulses of the experiments and the simulation are a good match (see  \creft{tab:VP_statistical}, \crefs{fig:GRFx}{b}{ and }{fig:GRFy}{b}). The normalized horizontal impulses for both braking and propulsion intervals are the same at $0.1$, while the normalized net vertical impulse in experiments are \SI{15}{\percent} higher than in simulation.
%
For the step-down conditions, the simulation predicts higher normalized net vertical impulse values of $1.46$ at step~0 and $1.36$ at step~1, as opposed to $1.31$ for the V10 condition and $1.18$ for C10 condition in experiments.
The change in the horizontal impulses during the step-down differs significantly between the simulation and experiments.
The V10 condition shows no significant change in the horizontal impulses, while in the C10 condition they decrease to $0.04$ for breaking and $0.06$ for propulsion.
%
In contrast, the simulations show an increase in the horizontal impulses (\creff{fig:GRFx}{b}). In particular for a step-down perturbation of \SI{\minus 10}{\centi\meter}, the normalized braking impulse increases to $0.15$ at step~0 and $0.18$ at step~1, whereas for propulsion it increases to $0.15$ and  $0.12$.

%
%

The different behavior we observe in horizontal impulses at step-down for the experiment and simulations may be due to different leg angles at touch-down.
We expect that a steeper leg angle of attack at touch-down would decrease the horizontal and increase the vertical braking impulse.
However, we observe with \SI{66}{\degree} a \SI{9}{\degree} steeper angle of attack in the simulations for level running than it was reported for V0 for the same experiments \cite{muller2012leg}.
Nevertheless, no corresponding changes in the braking impulses could be observed.
On the other hand, in the perturbed condition the angle of attack is with \SI{66}{\degree} nearly the same in the simulation and C10, but here the braking impulses differ.
Therefore, we conclude that additional factors have to be involved in the explanation of the different impulses between simulation and experiments and further investigations are needed.
The simulation could potentially be improved by implementing a swing-leg retraction as observed in humans \cite{Seyfarth_2003,Blum_2010,muller2012leg}.



\begin{figure}[!t]
{\captionof{figure}{Kinetic energy of the CoM for the human running experiments\protect\footnotemark[\value{footnote}] (left) and simulated model (right). The TSLIP model is able to predict the kinetic energies for the unperturbed and visible perturbed step well.  The simulation yields larger energy fluctuations during the stance phase compared to experiments. Experiments with camouflaged perturbation (C10) yield higher mean kinetic energy compared to the ones with visible perturbations (V10).
}\label{fig:ExpVsModel_Ekin}}
{\begin{annotatedFigure}
	{\includegraphics[width=1\linewidth]{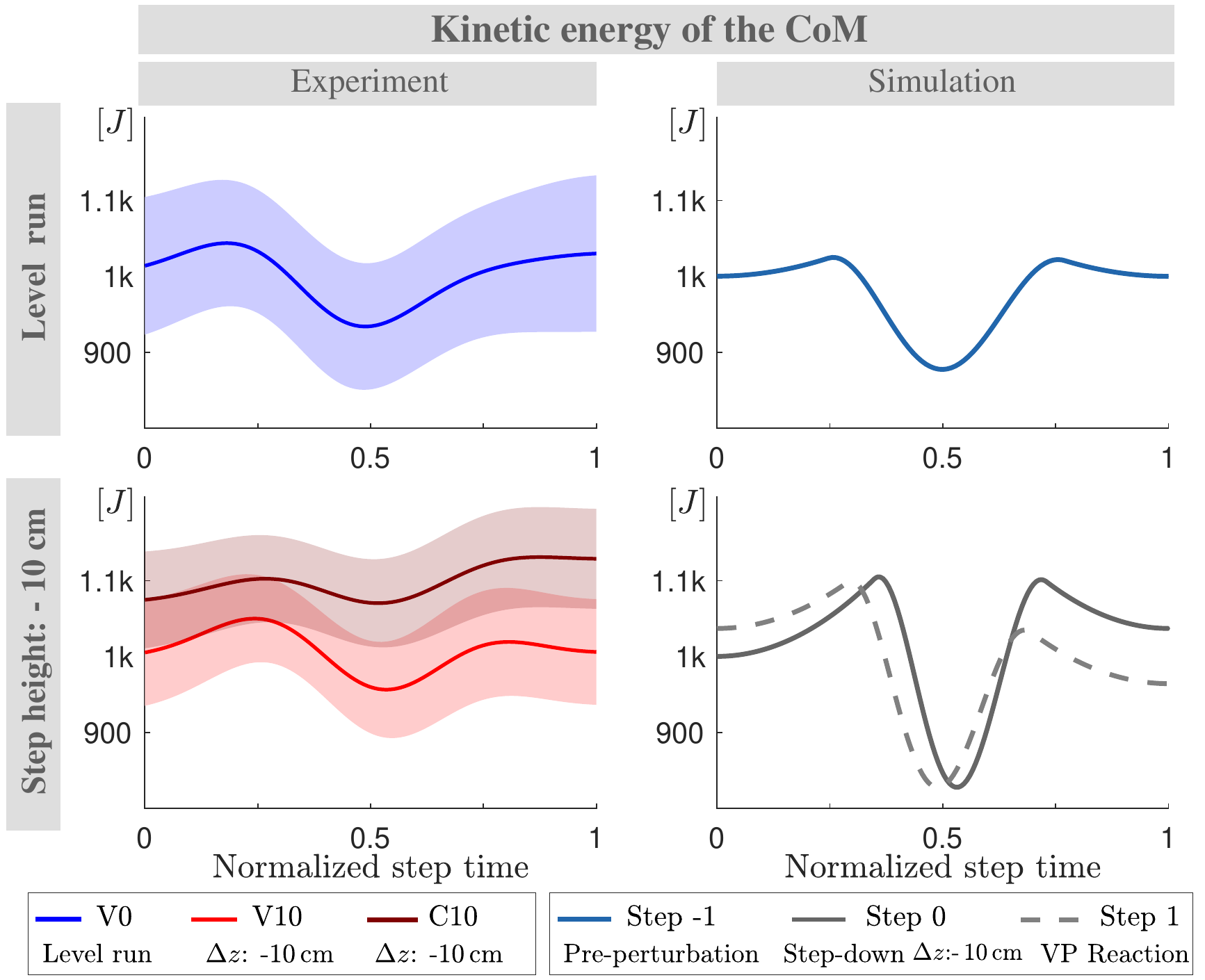}}
	\sublabel{a)}{0.15,0.87}{color_gray}{1}
	\sublabel{b)}{0.62,0.87}{color_gray}{1}
	\sublabel{c)}{0.15,0.48}{color_gray}{1}
	\sublabel{d)}{0.62,0.48}{color_gray}{1}
\end{annotatedFigure}}
\end{figure}

In terms of the CoM energies, there is a good match between the kinetic energies of the experiments and simulation for the unperturbed step (V0 and step~{$\minus 1$} in \crefsub{fig:ExpVsModel_Ekin}{a}{b}). 
The simulated energies of the perturbed step are closer to the experiments with visible perturbations (V10 and steps~0~and~1 in \crefsub{fig:ExpVsModel_Ekin}{c}{d}).
Human experiments show a drop in kinetic energy of \SI{9}{\percent} for V10, \SI{3}{\percent} for C10.
The simulation shows a drop in kinetic energy of about \SI{25}{\percent} for step~0 and step~1.
The C10 condition shows a higher mean kinetic energy compared to visible perturbations and there is no obvious decrease of energy in the stance phase (\creff{fig:ExpVsModel_Ekin}{c}) .

\begin{figure}[t!]
{\captionof{figure}{Potential energy of the CoM  for the human running experiments\protect\footnotemark[\value{footnote}] (left) and simulated model (right). Overall, the TSLIP model predicts the CoM height and its related potential energy well.
}\label{fig:ExpVsModel_Epot}}
{\begin{annotatedFigure}
	{\includegraphics[width=1\linewidth]{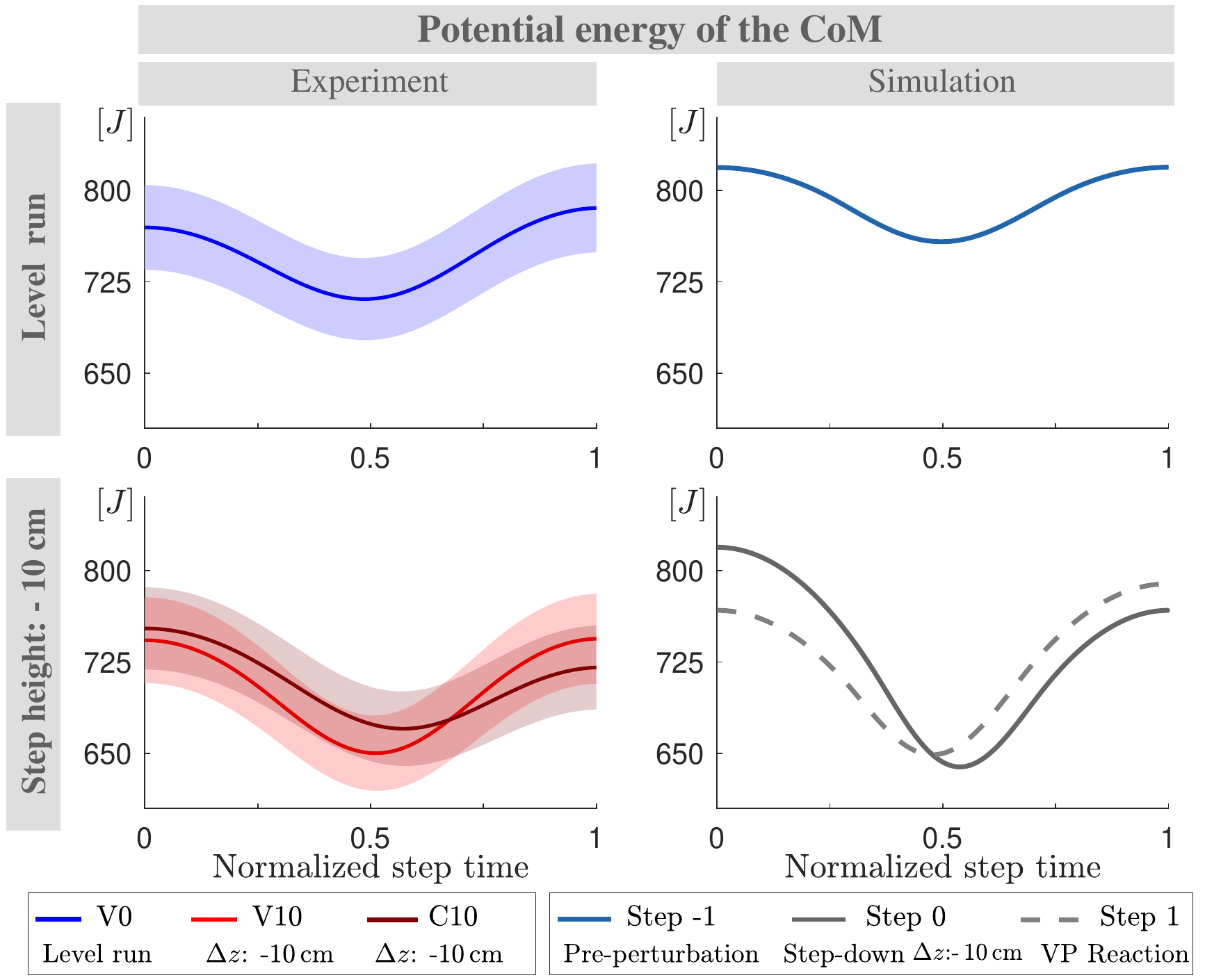}}
	\sublabel{a)}{0.15,0.87}{color_gray}{1}
	\sublabel{b)}{0.62,0.87}{color_gray}{1}
	\sublabel{c)}{0.15,0.48}{color_gray}{1}
	\sublabel{d)}{0.62,0.48}{color_gray}{1}
\end{annotatedFigure}}
\end{figure}

%
 The potential energy estimate of the simulations lies in the upper boundary of the experiments for the unperturbed step (V0 and step~-1 in \crefsub{fig:ExpVsModel_Ekin}{a}{b}).
The experiments with visible and camouflaged perturbations, as well as the TSLIP model, result in similar potential energy curves (\crefsub{fig:ExpVsModel_Ekin}{c}{d}).

\subsection{Limitations of this study}\label{subsec:Limitations}
The human experiments and the numerical simulations differ in several points, and conclusions from a direct comparison must be evaluated carefully. We discuss details for our choice of human experimental and numerical simulation conditions in this section.

First of all, there is a difference in terrain structure. After passing step~0, the human subjects face a different terrain structure type, compared to the TSLIP simulation model. The experimental setup is constructed as a pothole: a step-down followed by a step-up. However, an identical step-up in the numerical simulation would require an additional set of controllers to adjust the TSLIP model's leg angle and push off energy. Hence for the sake of simplicity, the TSLIP model continues running on the lower level and without a step-up. After the step-down perturbation, the simulated TSLIP requires several steps to recover. An experimental setup for an equivalent human experiment would require a large number of force plates, which were not available here.

In the V10 condition, the subjects have a visual feedback and hence the prior knowledge of the upcoming perturbation. This additional information might affect the chosen control strategy.
In particular, since there is a step-up in the human experiments, subjects might account for this upcoming challenge prior to the actual perturbation.
%

In the C10 condition, some subjects might prioritize safety in the case of a sudden and expected drop, and employ additional reactive strategies \cite{muller2015preparing}.
In contrast, the simulations with a VP controller can not react to changes during the step-down and only consider the changes of the previous step when planning for the next.

Furthermore, in the human experiments we can not set a step-down higher than \SI{\minus 10}{\centi\meter} due to safety reasons, especially in the camouflaged setting.
Instead, we can evaluate these situations in numerical simulations and test whether a hypothesized control mechanism can cope with higher perturbations.
However, one has to keep in mind that the TSLIP model that we utilize in our analysis is simplified. Its single-body assumption considers neither intra-segment interactions, nor leg dynamics from impacts and leg swing. Finally, our locomotion controller applied does not mimic specific human neural locomotion control or sensory feedback strategy.

%% file: sec/conclusion.tex
In this work, we investigate the existence and position of a virtual point (VP) in human running gait, and analyzed the implications of the observed VP location to postural stability and energetics with the help of a numerical simulation.

In addition to level running, we also inquired into the change of VP position when stepping down on a \SI{\minus 10}{\centi\meter} visible or camouflaged drop.
Our novel results are two-fold: First, the ground reaction forces focus around a point that is \SI{\minus 30}{\centi\meter} below the center of mass (CoM) for the human running at \SI{5}{\meter\per\second}.
The VP position does not change significantly when stepping down a visible or camouflaged drop of \SI{\minus 10}{\centi\meter}.
Second, the TSLIP model simulations show that a VP target below the center of mass (\VPb) is able to stabilize the body against step-down perturbations without any need to alter the state or model parameters.

%% file: sec/acknowledgement.tex

The authors thank the International Max Planck Research School for Intelligent Systems (IMPRS-IS) for supporting {\"O}zge Drama. This work was partially made thanks to a Max Planck Group Leader grant awarded to A.~Badri-Spr{\"o}witz by the Max Planck Society.
We also thank Martin G{\"o}tze and Michael Ernst for supporting the experiments.
The human running project was supported by the German Research Foundation (Bl 236/21 to Reinhard Blickhan and MU 2970/4-1 to Roy M{\"u}ller).

%% file: sec/data_availability.tex
Kinetic and kinematic data of the human running experiments are available from the figshare repository: \url{https://figshare.com/s/f52bb001615718f5de80}

%% file: sec/nomenclature.tex
\nomenclature[A, 001]{CoM}{Center of mass}
\nomenclature[A, 002]{TSLIP}{Spring loaded inverted pendulum model extended with a trunk}
\nomenclature[A, 003]{VP}{Virtual point}
\nomenclature[A, 004]{\VPa}{Virtual point above the center of mass}
\nomenclature[A, 005]{\VPb}{Virtual point below the center of mass}
\nomenclature[A, 006]{\VPbl}{Virtual point below the center of mass and below the leg axis at touch-down}
\nomenclature[A, 007]{$g$}{$g=9.81$\si{\metre\squared\per\second},~Standard acceleration due to gravity}

\nomenclature[B, 001]{CoP}{Center of pressure}
\nomenclature[B, 002]{GRFs}{Ground reaction forces}
\nomenclature[B, 003]{V0}{Experiment with level ground}
\nomenclature[B, 004]{V10}{Experiment with \SI{10}{\centi\meter} visible step-down perturbation}
\nomenclature[B, 005]{C10}{Experiment with \SI{10}{\centi\meter}  camouflaged step-down perturbation}
\nomenclature[B, 006]{$R^2$}{Coefficent of determination}
\nomenclature[B, 007]{$\gamma$}{The trunk angle estimated from markers on L5 and C7. The trunk angle $\gamma$ corresponds to the $\theta_{{C}}$ in the TSLIP model.}
\nomenclature[B, 001]{$l$}{Distance between lateral malleolus and trochanter major of the leg in contact with the ground}
\nomenclature[B, 008]{$N_{trial}$}{Number of trials}
\nomenclature[B, 009]{$N_{\%}$}{Number of gait percentage times analyzed}
\nomenclature[B, 010]{$\theta_{exp}$}{Angle of the experimental measured GRFs}
\nomenclature[B, 011]{$\overline{\theta}_{exp}$}{Mean experimental angle of GRFs}
\nomenclature[B, 012]{$\theta_{theo}$}{Angle of theoretical forces}
\nomenclature[B, 013]{$\vec{p}$}{Impulse}
\nomenclature[B, 014]{$\vec{p}_{normalized}$}{Normalized impulse}
\nomenclature[B, 015]{$\vec{p}_{brake}$}{Braking impulse}
\nomenclature[B, 016]{$\vec{p}_{prop}$}{Propulsion impulse}
\nomenclature[C, 001]{$[x_{\protect{C}}, z_{\protect{C}}, \theta_{\protect{C}}]$}{State vector of the center of mass}
\nomenclature[C, 002]{$[r_{FC}, r_{FV},r_{FH}]$}{Position vectors from foot to the center of mass, virtual point and hip joint, respectively}
\nomenclature[C, 003]{$ \Delta z$}{Step-down height}
\nomenclature[C, 004]{$m$}{Mass}
\nomenclature[C, 005]{$J$}{Moment of inertia}
\nomenclature[C, 006]{$l$}{Leg length}
\nomenclature[C, 007]{$\theta_{L}$}{Leg angle}
%
\nomenclature[C, 009]{$\tau_{H}$}{Hip torque}
\nomenclature[C, 010]{$F_{sp}$}{Leg spring force}
\nomenclature[C, 011]{$F_{dp}$}{Leg damper force}
\nomenclature[C, 012]{$\prescript{}{F}{\mathbf{F}}_{a}$}{Axial component of the ground reaction force in foot frame}
\nomenclature[C, 013]{$\prescript{}{F}{\mathbf{F}}_{t}$}{Tangential component of the ground reaction force in foot frame}
%
\nomenclature[C, 014]{$r_{VP}$}{VP radius, the distance between the center of mass and virtual point}
\nomenclature[C, 015]{$\theta_{VP}$}{VP angle, the angle between trunk axis and \VPa, or the vertical axis passing from CoM and \VPb}
%

\nomenclature[D, 001]{AP}{Apex event, where the center of mass reaches to its max. height}
\nomenclature[D, 002]{TD}{Leg touch-down event}
\nomenclature[D, 003]{TO}{Leg take-off event}
\nomenclature[D, 004]{Des}{Desired value of the variable}

\nomenclature[E, 001]{$i$}{Current step}
\nomenclature[E, 002]{$i \protect\minus 1$}{Previous step}
%
\printnomenclature

%% file: sec/appendix.tex
\subsection{Simulation: TSLIP model parameters}\label{sec:app:simTSLIPparameters}

The TSLIP model parameters are presented in \creft{tab:ModelPrm}{} (see \cite{drama2019human} for the parameters for the human model and \cite{drama2019bird} for the avian model).
\begin{table}[h!]
\centering
\captionsetup{justification=centering}
\caption{Model parameters for TSLIP model}
\label{tab:ModelPrm}
\begin{adjustbox}{width=1\linewidth}
\begin{tabular}{@{} l| c c c c c  @{}}
\multicolumn{1}{l}{Name} & \multicolumn{1}{c}{Symbol} &  \multicolumn{1}{c}{Units} &   \multicolumn{1}{c}{Literature} & \multicolumn{1}{c}{Chosen} & \multicolumn{1}{c}{Reference}  \\
\hline
\hspace{1mm} mass & $\mathit{m}$  & \si{\kilogram}    &  60-80   & 80 &   {\cite{Sharbafi_2013}}  \\
\hspace{1mm} moment of inertia & $\mathit{J}$    & \si{\kilogram\meter\squared} & 5 & 5 &  {\cite{ Sharbafi_2013, deLeva_1996}}   \\
\hspace{1mm} leg stiffness & $\mathit{k}$  & \si{\kilo\newton\per\meter} &16-26 & 18 &  {\cite{Sharbafi_2013, McMahon_1990}}   \\
\hspace{1mm} leg length & $\mathit{l}$  & \si{\meter}   &  1  &  1 &  {\cite{Sharbafi_2013}}  \\
\hspace{1mm} leg angle at TD & $ \mathit{\theta_{L}^{TD}}$ & (\si{\degree}) & 78-71 & $\mathit{f_{H}(\dot{x})}$ & {\cite{Sharbafi_2013, McMahon_1990}}  \\
\hspace{1mm} dist. Hip-CoM & $\mathit{r_{HC}}$  & \si{\meter} & 0.1 &0.1 & {\cite{Sharbafi_2013,Wojtusch_2015}}
\end{tabular}
\end{adjustbox}
\vspace{-4mm}
\end{table}

\subsection{Simulation: Flowchart for leg angle and VP angle control}\label{sec:app:simTSLIPparameters}

\begin{figure}[tbh!]
{\captionof{figure}{ The linear feedback control scheme for the leg angle in \crefe{eqn:thetaL} and the VP angle in \crefe{eqn:thetaVP} are presented. Both controllers update step-to-step at the apex event where the CoM height reaches to its maximum.
}\label{fig:Flowchart}}
{\includegraphics[width=1\linewidth]{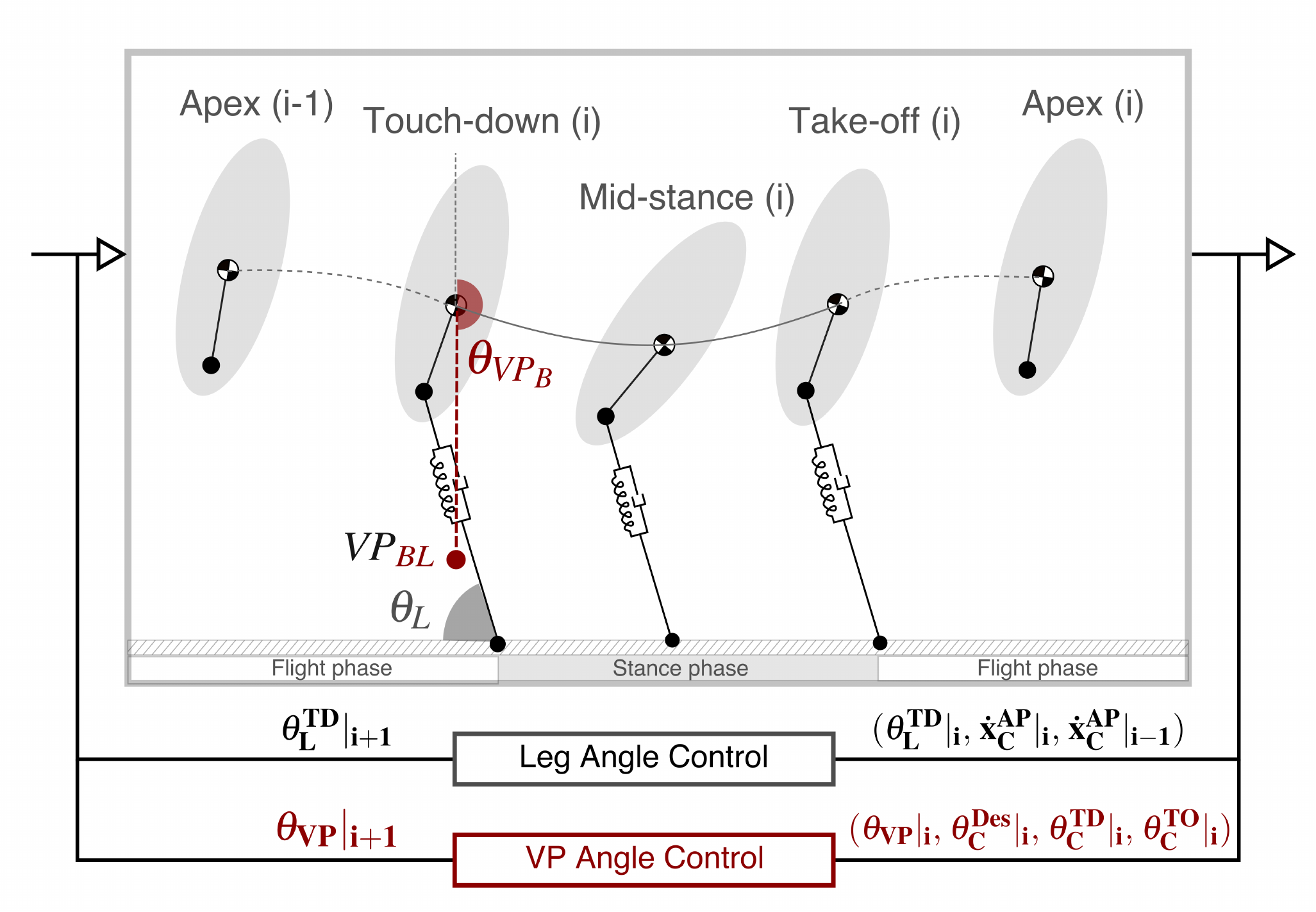}}
\end{figure}

The linear controller for the leg angle $\theta_{L}$ and VP angle $\theta_{VP}$ is presented in \creff{fig:Flowchart}{}. The leg angle control coefficients ($k_{\dot{x}} \, k_{\dot{x}_{0}}$) in \crefe{eqn:thetaL} are decreased from ($0.25, \,0.5  k_{\dot{x}}$) to ($0.2, \, 0.3 k_{\dot{x}}$), as the step-down height is increased from \SI{\minus 10}{\centi\meter} to  \SI{\minus 40}{\centi\meter}. The reduction of the coefficients slows down the adjustment of the forward speed, and enables us to prioritize the postural correction in the presence of larger perturbations.

%

\vfill\break

\subsection{Simulation: Energy regulation at the leg and hip}\label{sec:app:simLegHipEnergies}

Here, we present the energy levels of the leg spring, leg damper and the hip actuator for the entire set of step-down perturbations ($\Delta z \myeq$[\SI{ \minus 10, \minus 20,  \minus 30, \minus 40}{\centi\meter}]).

\begin{figure}[tbh!]
{\captionof{figure}{ The energy curves for the leg spring ($a_{0} \protect \minus a_{4}$), leg damper ($b_{0}  \protect \minus b_{4}$) and hip actuator ($c_{0}  \protect \minus c_{4}$){\protect\footnotemark}. The sub-index "0" indicates the trajectory belongs to the equilibrium state.  With the increase of the system's energy at step-down ({\protect\markerLine[color_darkgray]}), the leg deflects more, the leg damper dissipates more energy and the hip actuator injects more energy than its equilibrium condition. During the reaction step ({\protect\markerDashedLine[color_darkgray]}), the hip actuator reacts to energy change and starts to remove energy from the system. In the following steps ({\protect\markerLine[color_gray]}) the hip regulates the energy until the system reaches to the initial equilibrium state ({\protect\markerLine[color_blue]}).
}\label{fig:Energy_dt_Dz}}
{\begin{annotatedFigure}
	{\includegraphics[width=1\linewidth]{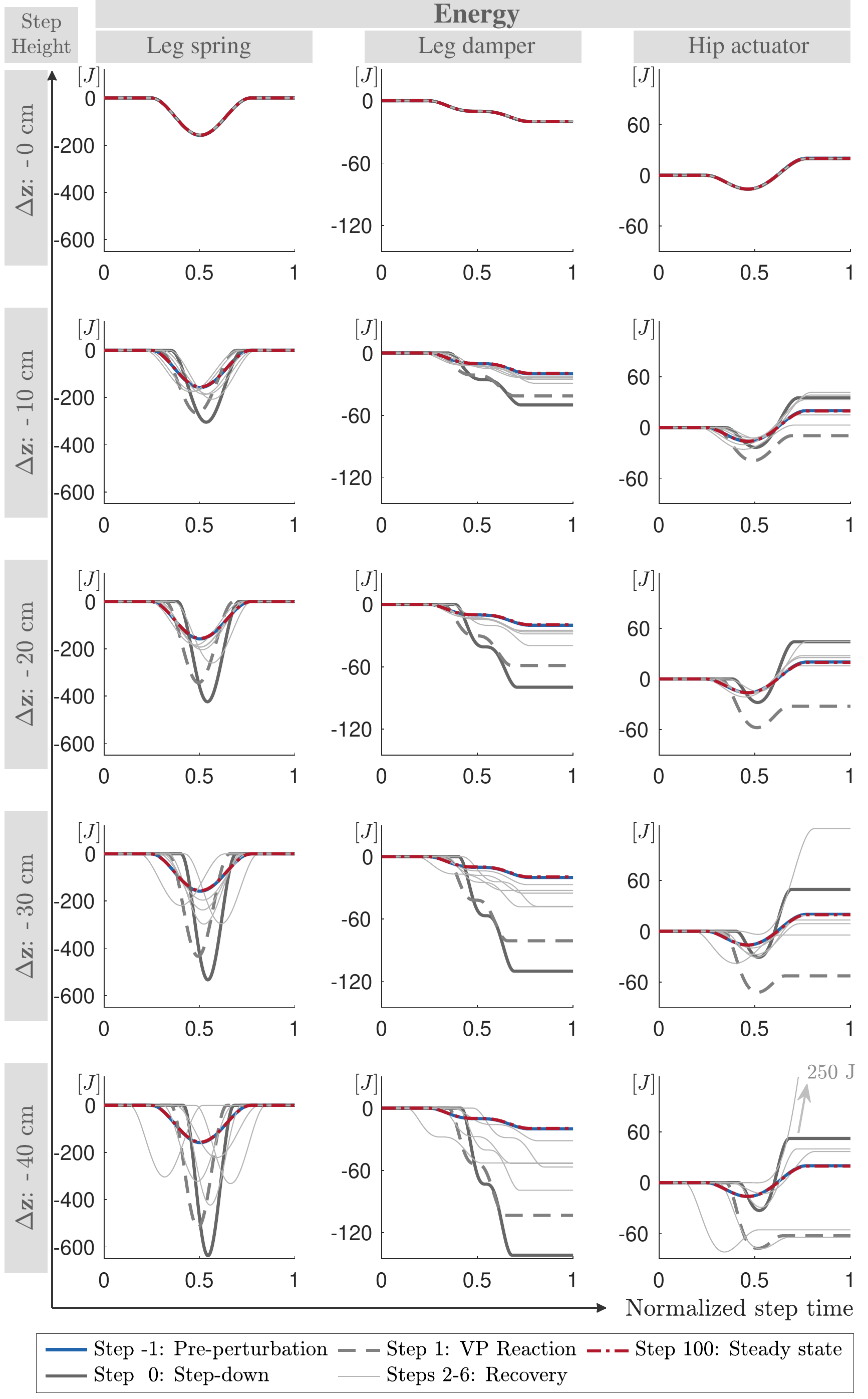}}
	\sublabel{$a_{0}$)}{0.15,0.945}{color_gray}{0.7}  \sublabel{$b_{0}$)}{0.47,0.945}{color_gray}{0.7}  \sublabel{$c_{0}$)}{0.79,0.945}{color_gray}{0.7}
	\sublabel{$a_{1}$)}{0.15,0.765}{color_gray}{0.7}     \sublabel{$b_{1}$)}{0.47,0.765}{color_gray}{0.7}    	\sublabel{$c_{1}$)}{0.79,0.765}{color_gray}{0.7}
	\sublabel{$a_{2}$)}{0.15,0.585}{color_gray}{0.7}  \sublabel{$b_{2}$)}{0.47,0.585}{color_gray}{0.7}	\sublabel{$c_{2}$)}{0.79,0.585}{color_gray}{0.7}
	\sublabel{$a_{3}$)}{0.15,0.405}{color_gray}{0.7}     \sublabel{$b_{3}$)}{0.47,0.405}{color_gray}{0.7}		\sublabel{$c_{3}$)}{0.79,0.405}{color_gray}{0.7}
	\sublabel{$a_{4}$)}{0.15,0.225}{color_gray}{0.7}     \sublabel{$b_{4}$)}{0.47,0.225}{color_gray}{0.7}	\sublabel{$c_{4}$)}{0.79,0.225}{color_gray}{0.7}
\end{annotatedFigure}}
\end{figure}
\footnotetext{In subplot $c_{4}$, the maximum value of steps 7-11 is indicated with a text and arrow due to the scaling issues.}

\clearpage\newpage

\subsection{Simulation: Ground reaction forces and impulses}\label{sec:app:simGRF}

We provide the vertical and horizontal ground reaction forces for the entire set of step-down perturbations ($\Delta z \myeq$[\SI{ \minus 10, \minus 20,  \minus 30, \minus 40}{\centi\meter}]).

\begin{figure}[hb!]
{\begin{annotatedFigure}
{\includegraphics[width=1\linewidth]{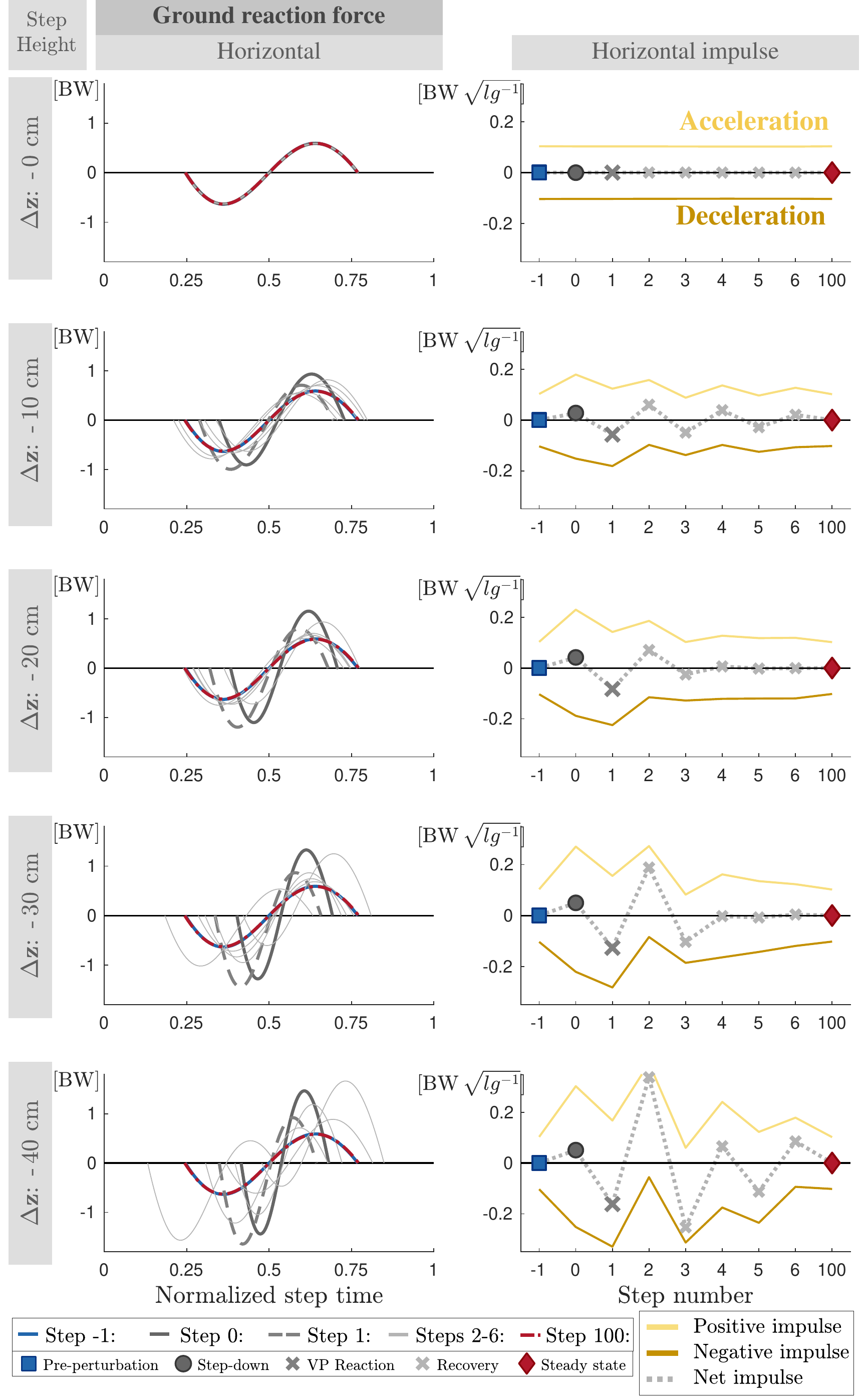}}
\sublabel{$a_{0}$)}{0.16,0.935}{color_gray}{0.7}  \sublabel{$b_{0}$)}{0.64,0.935}{color_gray}{0.7}
\sublabel{$a_{1}$)}{0.16,0.76}{color_gray}{0.7}  \sublabel{$b_{1}$)}{0.64,0.76}{color_gray}{0.7}
\sublabel{$a_{2}$)}{0.16,0.585}{color_gray}{0.7}  \sublabel{$b_{2}$)}{0.64,0.585}{color_gray}{0.7}
\sublabel{$a_{3}$)}{0.16,0.41}{color_gray}{0.7}  \sublabel{$b_{3}$)}{0.64,0.41}{color_gray}{0.7}
\sublabel{$a_{4}$)}{0.16,0.23}{color_gray}{0.7}  \sublabel{$b_{4}$)}{0.64,0.23}{color_gray}{0.7}
\end{annotatedFigure}}
\captionof{figure}{The horizontal ground reaction forces over normalized step time are shown ($a_{0} \protect \minus a_{4}$). The peak horizontal GRF increases with the step-down perturbation. The area under this curve is the horizontal impulse, which corresponds to the acceleration and deceleration the main body ($b_{0} \protect \minus b_{4}$). The step-down perturbation at step~0 increases the energy of the system. The increase in energy influences the net horizontal impulse, as the impulse attains a positive value ({\protect\markerCircle[color_darkgray]}) and causes the body to accelerate forward. In response, the VP position changes to create net negative impulse in the following step (i.e., step~1, {\protect \raisebox{-0.6 pt}{\markerCross[color_darkgray][1.5]}}) and decelerates the body. The VP position is adjusted until all the excess energy is removed from the system\,({\protect \markerCross[color_gray][1]}) and the gait reaches to an equilibrium state\,({\protect \raisebox{-0.3 pt}{\markerDiamond[color_red][0]}}).
}\label{fig:GRFx}
\end{figure}

\begin{figure}[t!]
{\begin{annotatedFigure}
{\includegraphics[width=1\linewidth]{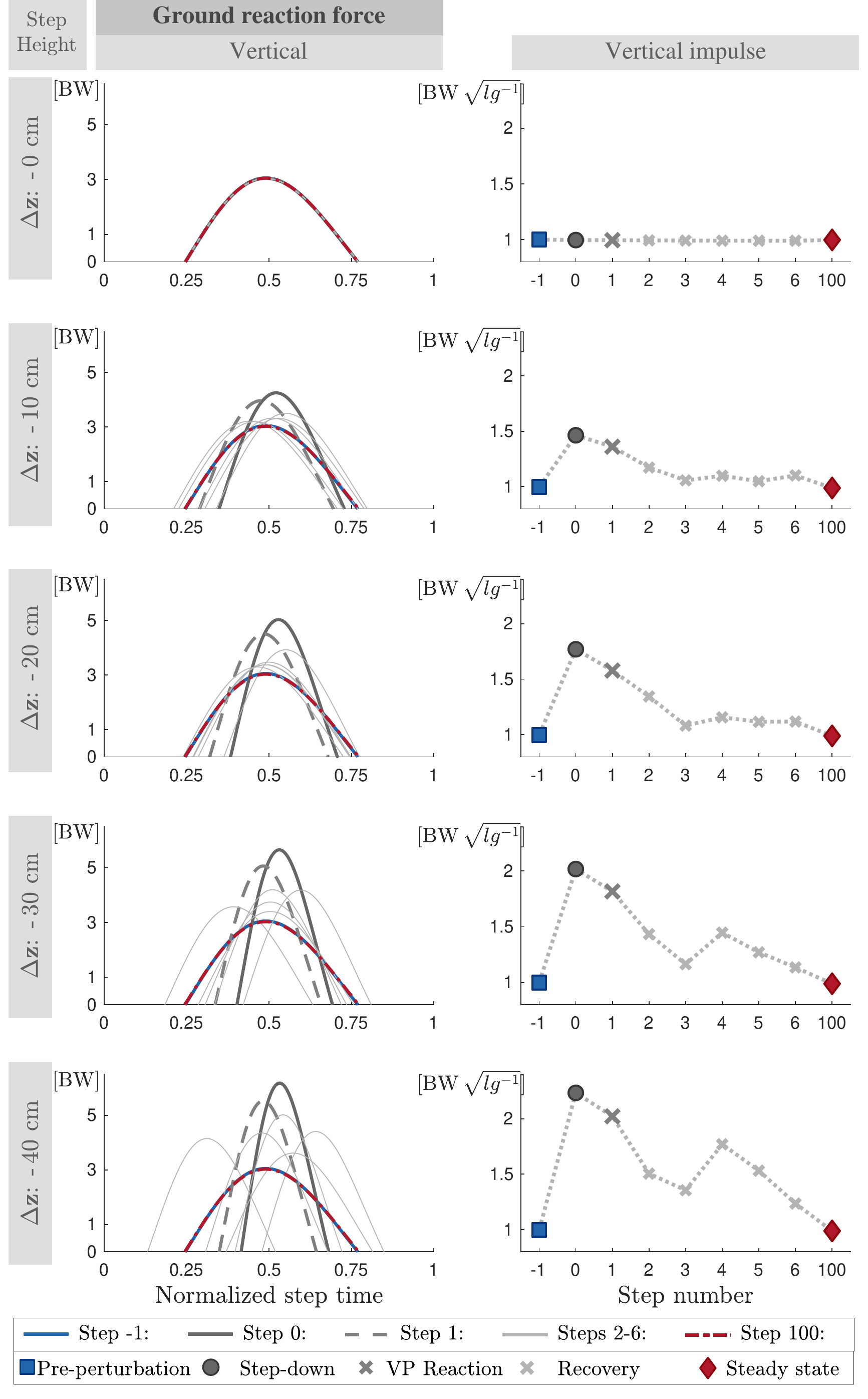}}
\sublabel{$a_{0}$)}{0.16,0.935}{color_gray}{0.7}  \sublabel{$b_{0}$)}{0.64,0.935}{color_gray}{0.7}
\sublabel{$a_{1}$)}{0.16,0.76}{color_gray}{0.7}  \sublabel{$b_{1}$)}{0.64,0.76}{color_gray}{0.7}
\sublabel{$a_{2}$)}{0.16,0.585}{color_gray}{0.7}  \sublabel{$b_{2}$)}{0.64,0.585}{color_gray}{0.7}
\sublabel{$a_{3}$)}{0.16,0.41}{color_gray}{0.7}  \sublabel{$b_{3}$)}{0.64,0.41}{color_gray}{0.7}
\sublabel{$a_{4}$)}{0.16,0.23}{color_gray}{0.7}  \sublabel{$b_{4}$)}{0.64,0.23}{color_gray}{0.7}
\end{annotatedFigure}}
\captionof{figure}{The vertical ground reaction forces over normalized step time are shown ($a_{0} \protect \minus a_{4}$). The peak vertical GRF increases with the step-down perturbation. During the following steps, the impulse decreases to its initial value through the regulation of the VP position. The increase in the peak GRF after step-down is proportional to the step-down height. Between steps 4-5, the peak vertical GRF increases 1.4 fold for \SI{\protect\minus 10}{cm} drop and 2 fold for \SI{\protect\minus 40}{cm} drop. In accordance, the vertical impulse increases with the step-down perturbation and returns to its initial value ($b_{0} \protect \minus b_{4}$). As the step-down height increases from -10 to \SI{\protect\minus 40}{cm} the vertical impulse increases 1.53 fold from its initial value for step~0\,({\protect \markerCircle[color_darkgray]}) and 1.48 fold for step~1\,({\protect \raisebox{-0.6 pt}{\markerCross[color_darkgray][1.5]}})
}\label{fig:GRFy}
\end{figure}

\clearpage \newpage

\subsection{STD of the Experiments}\label{sec:app:expSTD}
In the \cref{sec:discussion}, we provided the standard error (SE) of the measurements from the human running experiments (see the patched areas in \cref{fig:ExpVsModel_States,fig:ExpVsModel_Ekin,fig:ExpVsModel_Epot}).
The standard error is calculated by dividing the standard deviation by the square roof of number of subjects. The SE shows how good the mean estimate of the measurements is.

\begin{figure}[tbh!]
 \begin{floatrow}
  \ffigbox[\Xhsize]
{\begin{annotatedFigure}
{\includegraphics[width=0.97\linewidth]{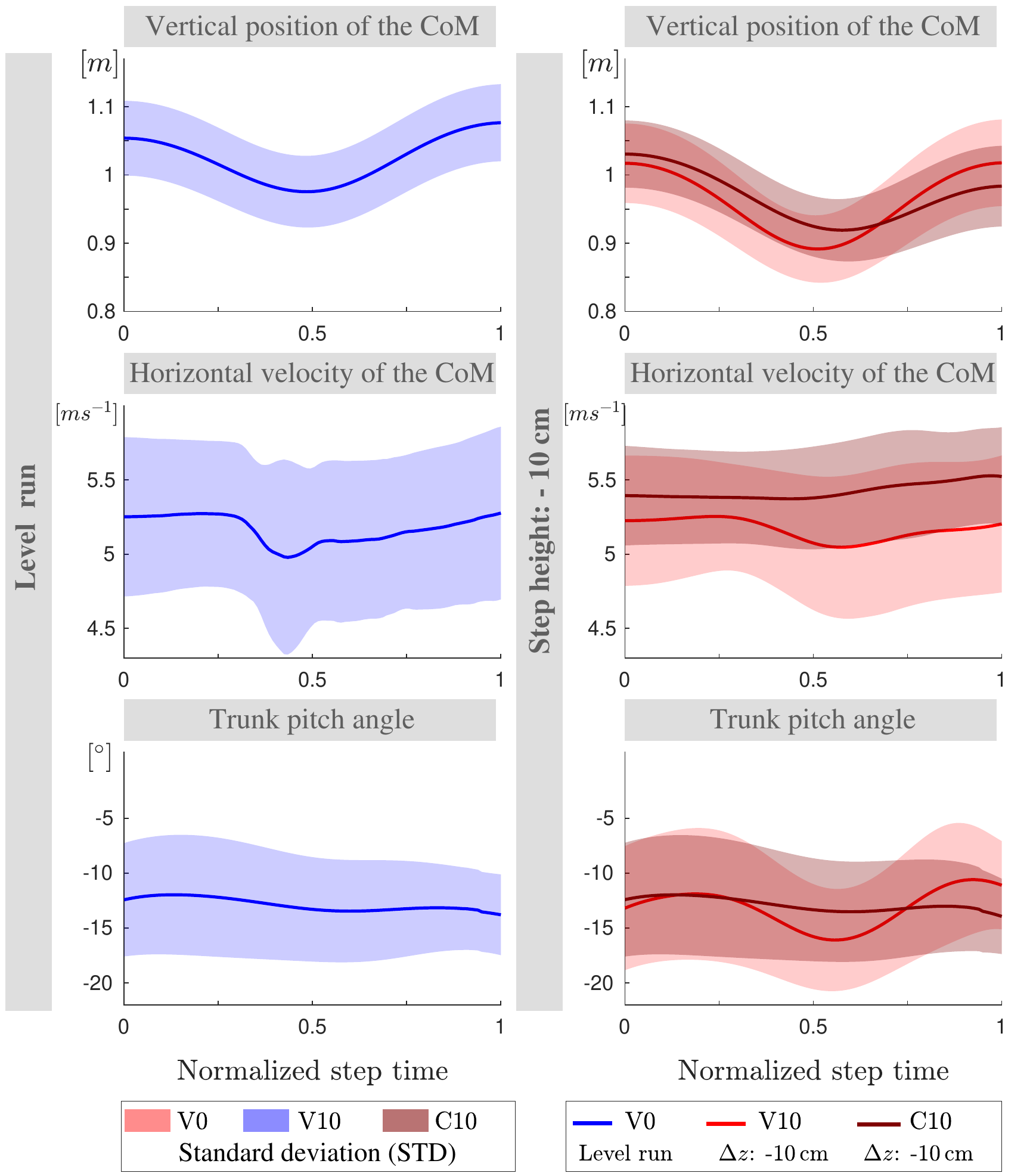}}
\sublabel{$a_{0}$)}{0.15,0.94}{color_gray}{0.7}
\sublabel{$a_{1}$)}{0.64,0.94}{color_gray}{0.7}
\sublabel{$b_{0}$)}{0.15,0.645}{color_gray}{0.7}
\sublabel{$b_{1}$)}{0.64,0.645}{color_gray}{0.7}
\sublabel{$c_{0}$)}{0.15,0.35}{color_gray}{0.7}
\sublabel{$c_{1}$)}{0.64,0.35}{color_gray}{0.7}
\end{annotatedFigure}}
{\caption{The figure is an extension of the \cref{fig:ExpVsModel_States}, with the difference that the standard deviation is plotted with the patches instead of the standard error.
}\label{fig:CoMstateSTD}}
 \end{floatrow}
  \vspace{\floatsep}%
  \begin{floatrow}
  \ffigbox[\Xhsize]
{\begin{annotatedFigure}
{\includegraphics[width=0.97\linewidth]{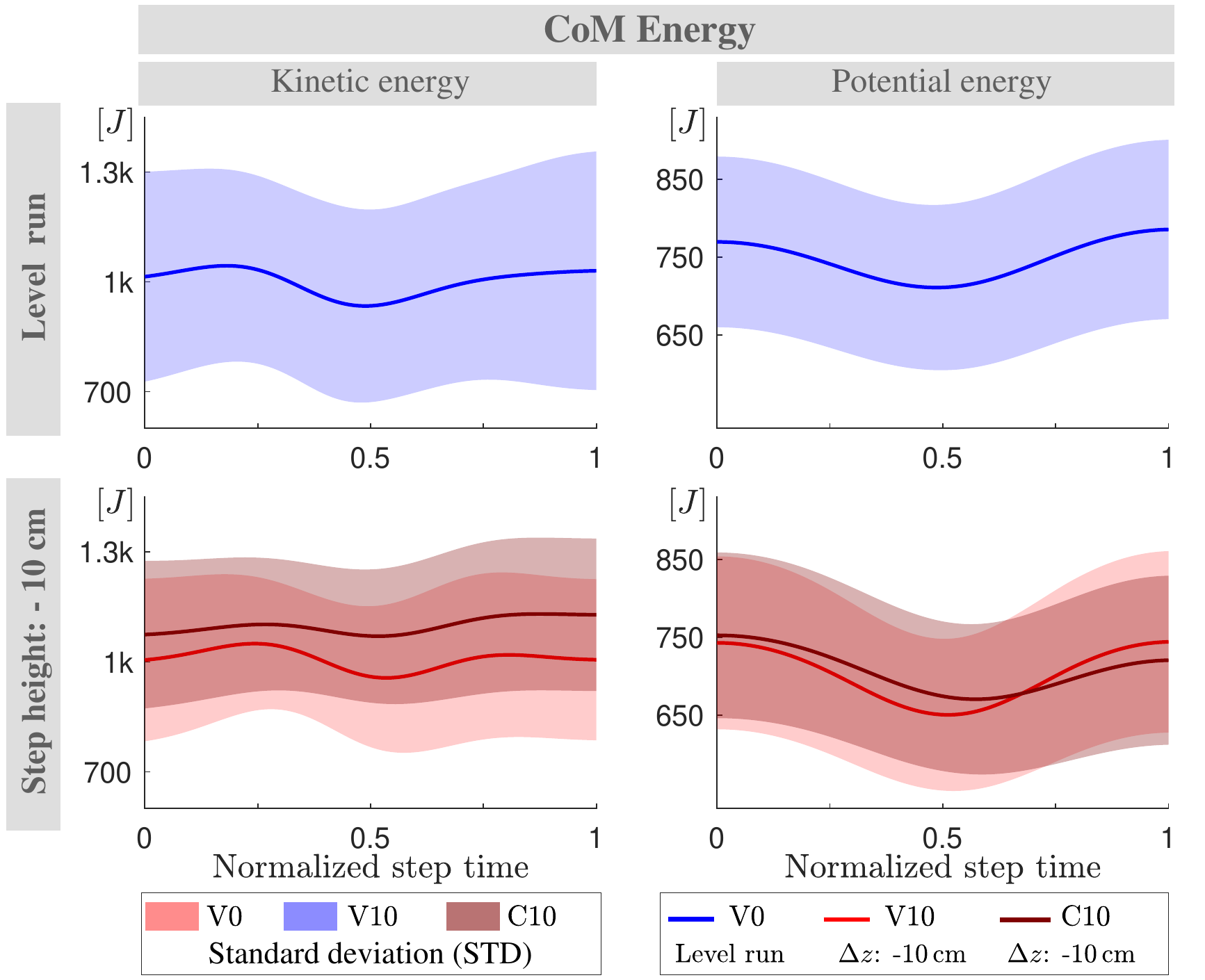}}
\sublabel{$a_{0}$)}{0.155,0.87}{color_gray}{0.7}
\sublabel{$b_{0}$)}{0.63,0.87}{color_gray}{0.7}
\sublabel{$a_{1}$)}{0.155,0.49}{color_gray}{0.7}
\sublabel{$b_{1}$)}{0.63,0.49}{color_gray}{0.7}
\end{annotatedFigure}}
{\caption{The figure is an extension of the \cref{fig:ExpVsModel_Ekin,fig:ExpVsModel_Epot}, with the difference that the standard deviation is plotted with the patches instead of the standard error.
}\label{fig:CoMenergySTD}}
\end{floatrow}
  \vspace{-2\floatsep}%
\end{figure}

On the other hand, standard deviation (STD) shows how spread out our different measurements are. The STD is an important measure, especially for the trunk angle measurements, where the trajectories of the each subject significantly varies. Therefore, we provide the STD values here in for the CoM state in \cref{fig:CoMstateSTD}  and CoM energy in \cref{fig:CoMenergySTD}.

\subsection{GRFs: Simulation vs. Experiment}\label{sec:app:simexpGRF}

\begin{figure}[tbh!]
{\begin{annotatedFigure}
{\includegraphics[width=1\linewidth]{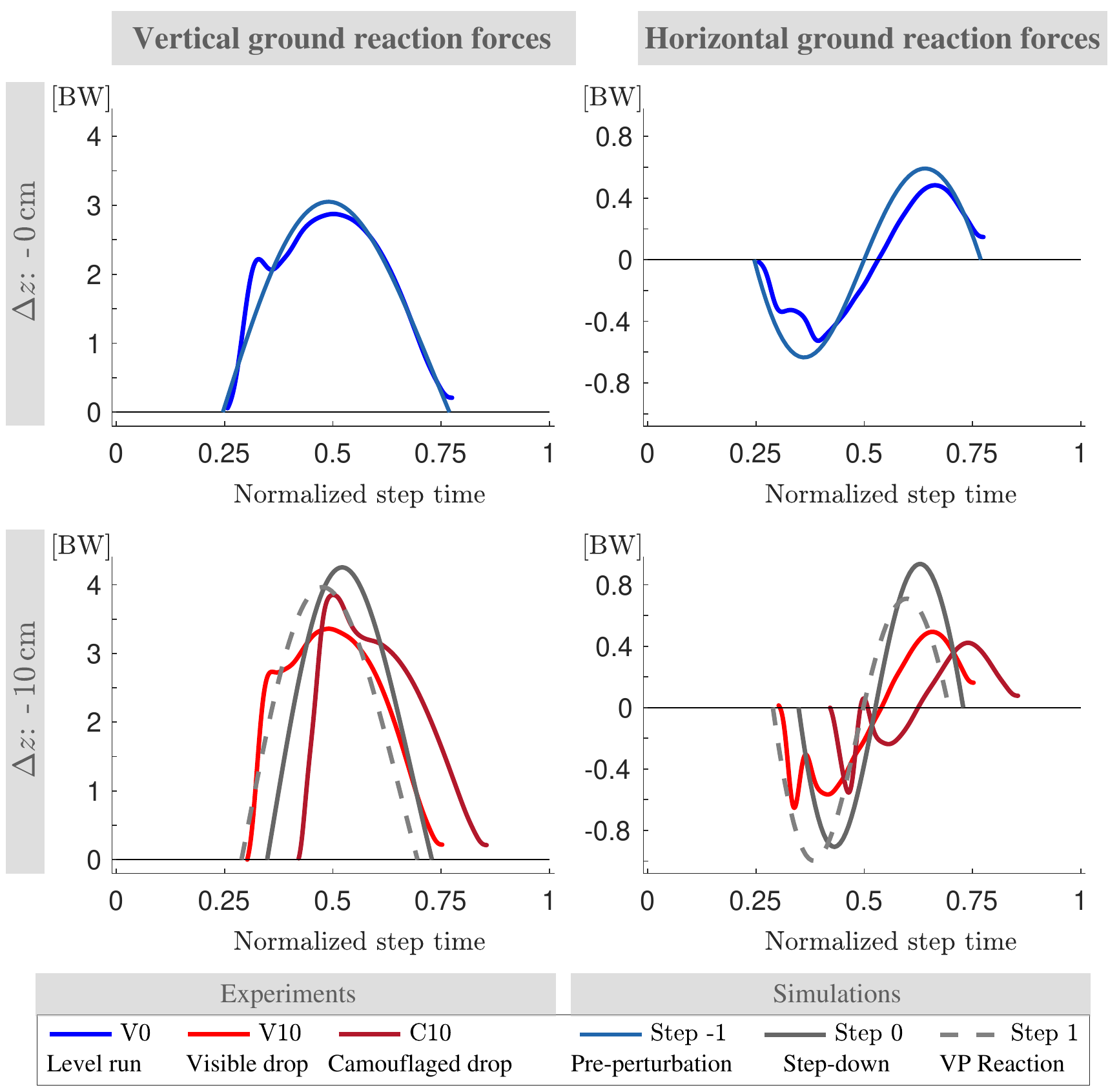}}
\sublabel{$a_{0}$)}{0.13,0.91}{color_gray}{0.7}
\sublabel{$b_{0}$)}{0.6,0.91}{color_gray}{0.7}
\sublabel{$a_{1}$)}{0.13,0.5}{color_gray}{0.7}
\sublabel{$b_{1}$)}{0.6,0.5}{color_gray}{0.7}
\end{annotatedFigure}}
\captionof{figure}{The vertical (a) and horizontal (b) ground reaction forces are plotted over normalized step time. The mean of the experimental results are shown. The TSLIP model simulation is able to capture the characteristics of the GRF  in level running  ($a_{0}$,$b_{0}$). For the step-down perturbation, the model predicts higher values for the peak vertical ($a_{1}$) and horizontal ($b_{1}$) GRF, compered to the mean values of the experiments.
}\label{fig:GRFexpsim}
\end{figure}

We present the vertical (a) and horizontal (b) GRFs belonging to the step~0 of the human running experiments (V0, V10, C10) and steps~-1,0 and 1 of the simulations with a \SI{\minus 10}{\centi\meter} step-down height, plotted on top of each other in \creff{fig:GRFexpsim}{}.

